\documentclass[11pt]{article}
\usepackage{chicagob}
\usepackage{latexsym}
\usepackage{eepic}
\usepackage{amssymb}

\newcommand{\worst}{\mb{w}}
\newcommand{\best}{\overline{\utf}}

\newcommand{\AAumann}{Anscombe and Aumann}
\newcommand{\AAu}{Anscombe-Aumann}

\newcommand{\ext}{\sqsupseteq}

\newcommand{\ssim}{\simeq}

\newcommand{\commentout}[1]{}

\newcommand{\uaicommentout}[1]{}
\newcommand{\ijcai}[1]{}

\newcommand{\prfcommentout}[1]{}

\newtheorem{THEOREM}{Theorem}[section]
\newenvironment{theorem}{\begin{THEOREM}}%
                        {\end{THEOREM}}
\newtheorem{LEMMA}[THEOREM]{Lemma}
\newenvironment{lemma}{\begin{LEMMA}}%
                      {\end{LEMMA}}
\newtheorem{COROLLARY}[THEOREM]{Corollary}
\newenvironment{corollary}{\begin{COROLLARY}}%
                          {\end{COROLLARY}}
\newtheorem{PROPOSITION}[THEOREM]{Proposition}
\newenvironment{proposition}{\begin{PROPOSITION}}%
                            {\end{PROPOSITION}}
\newtheorem{DEFINITION}[THEOREM]{Definition}
\newenvironment{definition}{\begin{DEFINITION}\rm}%
                            {\end{DEFINITION}}
\newtheorem{CLAIM}[THEOREM]{Claim}
\newenvironment{claim}{\begin{CLAIM}\rm}%
                            {\end{CLAIM}}
\newtheorem{EXAMPLE}[THEOREM]{Example}
\newenvironment{example}{\begin{EXAMPLE}\rm}%
                            {\end{EXAMPLE}}
\newtheorem{REMARK}[THEOREM]{Remark}
\newenvironment{remark}{\begin{REMARK}\rm}%
                            {\end{REMARK}}

                      {}

\newcommand{\thm}{\begin{theorem}}
\newcommand{\lem}{\begin{lemma}}
\newcommand{\pro}{\begin{proposition}}
\newcommand{\dfn}{\begin{definition}}
\newcommand{\rem}{\begin{remark}}
\newcommand{\xam}{\begin{example}}
\newcommand{\cor}{\begin{corollary}}
\newcommand{\prf}{\trivlist \item[\hskip \labelsep{\bf Proof:}]}
\newcommand{\ethm}{\end{theorem}}
\newcommand{\elem}{\end{lemma}}
\newcommand{\epro}{\end{proposition}}
\newcommand{\edfn}{\bbox\end{definition}}
\newcommand{\erem}{\bbox\end{remark}}
\newcommand{\exam}{\bbox\end{example}}
\newcommand{\ecor}{\end{corollary}}
\newcommand{\eprf}{\bbox\endtrivlist}
\newcommand{\beqn}{\begin{equation}}
\newcommand{\eeqn}{\end{equation}}

\newcommand{\beqna}{\begin{eqnarray}}
\newcommand{\eeqna}{\end{eqnarray}}

\newcommand{\bbox}{\vrule height7pt width4pt depth1pt}

\newcommand{\clm}{\begin{claim}}
\newcommand{\eclm}{\end{claim}}
\newcommand{\union}{\cup}
\newcommand{\sedom}{\mbb{E}}

\newcommand{\IR}{\realNum}

\renewcommand{\phi}{\varphi}

\newcommand{\E}{{\cal E}}

\renewcommand{\P}{{\cal P}}

\newcommand{\abs}[1]{\left|{#1}\right|}

\newcommand{\st}{\  \vert \ } %

\newcommand{\eg}{e.g.,}
\newcommand{\ie}{i.e.,}

\newcommand{\ol}{\setlength{\itemsep}{0pt}\begin{enumerate}}
\newcommand{\eol}{\end{enumerate}\setlength{\itemsep}{-\parsep}}
\newcommand{\ul}{\setlength{\itemsep}{0pt}\begin{itemize}}
\newcommand{\dl}{\setlength{\itemsep}{0pt}\begin{description}}
\newcommand{\edl}{\end{description}\setlength{\itemsep}{-\parsep}}
\newcommand{\eul}{\end{itemize}\setlength{\itemsep}{-\parsep}}

\newcommand{\mpro}[1]{{Proposition~\ref{#1}}}

\newcommand{\mthm}[1]{{Theorem~\ref{#1}}}

\newcommand{\msec}[1]{{Section~\ref{#1}}}

\newcommand{\mxam}[1]{{Example~\ref{#1}}}
\newcommand{\meqn}[1]{{(\ref{#1})}}

\newcommand{\blst}{\begin{list}{}{}}
\newcommand{\elst}{\end{list}}
\newcommand{\ben}{\begin{enumerate}}
\newcommand{\een}{\end{enumerate}}
\newcommand{\bit}{\begin{itemize}}
\newcommand{\eit}{\end{itemize}}
\newcommand{\bl}{\item}

\newcommand{\mc}[1]{\mathcal{#1}}
\newcommand{\mr}[1]{\mathrm{#1}}

\newcommand{\mb}[1]{\mathbf{#1}}

\newcommand{\mbb}[1]{\mathbb{#1}}

\newcommand{\paper}{paper}

\newcommand{\bfig}{\begin{figure}}
\newcommand{\efig}{\end{figure}}

\newcommand{\rar}{\rightarrow}

\newcommand{\urv}{utility random variable}
\newcommand{\urvs}{utility random variables}
\newcommand{\ulottery}{utility lottery}
\newcommand{\ulotteries}{utility lotteries}
\newcommand{\bydef}[1]{#1}

\newcommand{\ds}{\displaystyle}

\newcommand{\eset}{\emptyset}

\newcommand{\act}{a}
\newcommand{\sfa}{\act}
\newcommand{\actb}{b}
\newcommand{\stt}{s}

\newcommand{\csq}{c}
\newcommand{\csqd}{d}

\newcommand{\lto}{\ell}

\newcommand{\Act}{A}
\newcommand{\Stt}{S}

\newcommand{\Csq}{C}

\newcommand{\Lto}{L}

\newcommand{\HLT}{\mc{H}}
\newcommand{\Hlt}{H}
\newcommand{\hlt}{h}

\newcommand{\ACT}{\mc{A}}
\newcommand{\STT}{\mc{S}}

\newcommand{\LTO}{\mc{L}}

\newcommand{\ala}{\`{a}~la}

\renewcommand{\E}{\mb{E}}

\def\bbbu{{\mathchoice {\setbox0=\hbox{$\displaystyle\rm U$}\hbox{\hbox
to0pt{\kern0.4\wd0\vrule height1\ht0\hss}\box0}}
{\setbox0=\hbox{$\textstyle\rm U$}\hbox{\hbox
to0pt{\kern0.4\wd0\vrule height1\ht0\hss}\box0}}
{\setbox0=\hbox{$\scriptstyle\rm U$}\hbox{\hbox
to0pt{\kern0.4\wd0\vrule height1\ht0\hss}\box0}}
{\setbox0=\hbox{$\scriptscriptstyle\rm U$}\hbox{\hbox
to0pt{\kern0.4\wd0\vrule height1\ht0\hss}\box0}}}}

\def\bbbo{{\mathchoice {\setbox0=\hbox{$\displaystyle\rm O$}\hbox{\hbox
to0pt{\kern0.27\wd0\vrule height0.9\ht0\hss}\box0}}
{\setbox0=\hbox{$\textstyle\rm O$}\hbox{\hbox
to0pt{\kern0.27\wd0\vrule height0.9\ht0\hss}\box0}}
{\setbox0=\hbox{$\scriptstyle\rm O$}\hbox{\hbox
to0pt{\kern0.27\wd0\vrule height0.9\ht0\hss}\box0}}
{\setbox0=\hbox{$\scriptscriptstyle\rm O$}\hbox{\hbox
to0pt{\kern0.27\wd0\vrule height0.9\ht0\hss}\box0}}}}

\newcommand{\utf}{\mb{u}}

\newcommand{\regf}{\mb{r}}

\newcommand{\realNum}{\mbb{R}}

\newcommand{\ind}{\sim}

\newcommand{\ep}{\oplus}
\newcommand{\EP}{\Oplus}

\newcommand{\et}{\otimes}

\newcommand{\eleq}{\precsim}

\newcommand{\egr}{\succ}

\newcommand{\druletr}{\drule\ transformation}

\newcommand{\Oplus}{\bigoplus}

\newcommand{\bottom}{\bot}

\newcommand{\plf}{\mr{Pl}}
\newcommand{\Pl}{\mr{Pl}}

\newcommand{\ts}{\textstyle}

\newcommand{\plstic}{plausibilistic}

\newcommand{\nonplstic}{nonplausibilistic}

\newcommand{\edom}{E}

\newcommand{\leqact}{\precsim_{\Act}}

\newcommand{\pdom}{P}
\newcommand{\udom}{U}
\newcommand{\dom}{\mr{dom}}
\newcommand{\ran}{\mr{ran}}
\newcommand{\ssupp}{\mr{supp}}
\newcommand{\supp}{\mr{supp}}

\newcommand{\setthmnum}[2]{%
\setcounter{section}{#1}%
\setcounter{THEOREM}{#2}%
}

\newcounter{secStack}
\newcounter{thmStack}

\newcommand{\pushsec}{\setcounter{secStack}{\value{section}}}
\newcommand{\popsec}{\setcounter{section}{\value{secStack}}}
\newcommand{\pushthm}{\setcounter{thmStack}{\value{THEOREM}}}
\newcommand{\popthm}{\setcounter{THEOREM}{\value{thmStack}}}
\newcommand{\setThm}[2]{\pushsec\pushthm\renewcommand{\thesection}{\arabic{section}}\setthmnum{\value{#1}}{\value{#2}}}
\newcommand{\unsetThm}{\popsec\popthm\renewcommand{\thesection}{\Alph{section}}}

\newcommand{\trans}{\tau}

\newcommand{\congruent}{congruent}

\newcommand{\dr}{\mc{R}}
\newcommand{\dprb}{\mc{D}}

\newcommand{\pdprb}{\mc{D}}

\newcommand{\pldom}{\pdom}
\newcommand{\eudom}{V}

\newcommand{\pleq}{\preceq}

\newcommand{\leqRPdp}[2]{\precsim_{\mc{#1}({#2})}}

\newcommand{\leqRdp}[1]{\precsim_{\mc{R}({#1})}}

\newcommand{\drule}{decision rule}

\newcommand{\drules}{decision rules}

\newcommand{\DRules}{Decision Rules}

\newcommand{\Drule}{Decision rule}

\newcommand{\expstr}{expectation domain}

\newcommand{\expstrs}{expectation domains}

\newcommand{\ceu}{CEU}

\newcommand{\seu}{EU}
\newcommand{\gseu}{GEU}

\newcommand{\ndef}{{\uparrow}}

\newcommand{\maximin}{{Maxi\-min}}
\newcommand{\regret}{\mb{r}}

\newcommand{\reg}{{Mini\-max Regret}}

\newcommand{\Mmeu}{{Max\-min Expected Utility}}

\newcommand{\equm}{Expected Qualitative Utility Maximization}

\newcommand{\vonNM}{von~Neumann and Morgenstern}
\newcommand{\vNM}{vNM}

\newcommand{\npsub}{nonempty proper subset}

\newcommand{\represent}{represent}
\newcommand{\representation}{representation}
\newcommand{\representations}{representations}
\newcommand{\representable}{representable}

\newcommand{\represents}{represents}
\newcommand{\representing}{representing}
\newcommand{\represented}{represented}

\newcommand{\uniform}{uniform}

\newcommand{\nonuniform}{nonuniform}

\newcommand{\consequences}{consequences}

\newcommand{\ssofn}{states of the world}

\newcommand{\prefrel}{preference relation}
\newcommand{\prefrels}{preference relations}

\newcommand{\dprobs}{decision problems}

\newcommand{\dprob}{decision problem}
\newcommand{\dsitu}{decision situation}
\newcommand{\dsitus}{decision situations}

\newcommand{\sbset}{\subseteq}

\newcommand{\acts}{acts}

\newcommand{\tandb}{tastes and beliefs}
\newcommand{\dmakers}{DMs}
\newcommand{\dmaker}{DM}

\newcommand{\plstr}{plausibility domain}
\newcommand{\plstrs}{plausibility domains}
\newcommand{\plmsr}{plausibility measure}
\newcommand{\plmsrs}{plausibility measures}

\newcommand{\utstr}{utility domain}

\newcommand{\utstrs}{utility domains}
\newcommand{\utfn}{utility function}
\newcommand{\utfns}{utility functions}

\newcommand{\valstr}{valuation domain}

\newcommand{\Representing}{Representing}

\newcommand{\Bel}{\mr{Bel}}

\newcommand{\eprut}[1]{\E_{\Pr}(\utf_{#1})}
\newcommand{\eplut}[1]{\E_{\plf,\edom}(\utf_{#1})}

\newcommand{\ePplut}[2]{\E_{\plf,{#1}}(\utf_{#2})}

\newcommand{\deprut}[1]{\mbox{[[DELETE]]}}
\newcommand{\deplut}[1]{\mbox{[[DELETE]]}}

\newcommand{\vp}{\varphi}

\newcommand{\mmeu}{MMEU}

\newenvironment{oldthm}[1]{\par\noindent{\bf Theorem #1:} \em \noindent}{\par}
\newenvironment{oldlem}[1]{\par\noindent{\bf Lemma #1:} \em \noindent}{\par}
\newenvironment{oldcor}[1]{\par\noindent{\bf Corollary #1:} \em \noindent}{\par}
\newenvironment{oldpro}[1]{\par\noindent{\bf Proposition #1:} \em \noindent}{\par}
\newcommand{\othm}[1]{\begin{oldthm}{\ref{#1}}}
\newcommand{\eothm}{\end{oldthm} \medskip}
\newcommand{\olem}[1]{\begin{oldlem}{\ref{#1}}}
\newcommand{\eolem}{\end{oldlem} \medskip}
\newcommand{\ocor}[1]{\begin{oldcor}{\ref{#1}}}
\newcommand{\eocor}{\end{oldcor} \medskip}
\newcommand{\opro}[1]{\begin{oldpro}{\ref{#1}}}
\newcommand{\eopro}{\end{oldpro} \medskip}

\begin{document}

\title{Great Expectations.  
\\
Part II\@:  
Generalized Expected Utility
as a Universal Decision Rule%
\thanks{Work supported in part by NSF under grants IIS-0090145 and 
CTC-0208535 and by the DoD Multidisciplinary University Research
Initiative (MURI) program administered by ONR under
grant N00014-01-1-0795.
A preliminary version of this paper appeared in {\em Proceedings of
the 18th International Joint Conference on Artificial Intelligence
(IJCAI 2003), 2003}.}
}
\author{Francis C. Chu \and 
\commentout{
Joseph Y. Halpern\\
Department of Computer Science\\
Cornell University\\
Ithaca, NY 14853, U.S.A.\\
Email: \{fcc,halpern\}@cs.cornell.edu
}%
Joseph Y. Halpern}
\date{Department of Computer Science\\
Cornell University\\
Ithaca, NY 14853, U.S.A.\\
Email: \{fcc,halpern\}@cs.cornell.edu}

\maketitle

\begin{abstract}
Many different rules for decision making have been introduced in the
literature.  
We show that a notion of generalized expected utility
proposed in \cite{CH03a} is a universal
decision rule, in the sense that it can 
\represent\ 
essentially all other
decision rules.  
\end{abstract}

\section{Introduction}

A great deal of effort has been devoted to studying decision making.  
\commentout{
A standard 
formalization
assumes that the decision maker (DM) must choose
among 
}%
A standard formalization describes the choices 
a decision maker (DM) faces as 
acts, where an \emph{act}
is a function from states to
consequences.
Many \drules\ (that is, rules for choosing among acts, based
on the tastes and beliefs of the DM) have been proposed in the literature.
Some are meant to describe how ``rational'' agents should make decisions, 
while others aim at modeling how real agents actually make decisions.
Perhaps the best-known approach is that of 
\emph{maximizing expected utility}
(\seu)\@.   Normative arguments due to
Savage~\citeyear{Savage} suggest that 
rational agents
should 
behave
as if their tastes
are represented by a real-valued utility function 
on the consequences, their
beliefs about the likelihood of 
events (\ie\ sets of states) are represented by a
probability measure, and they are maximizing the expected utility of
acts
with respect to this utility and probability.

Despite these normative arguments,
it is well known that 
\commentout{
people do not 
always 
behave 
like expected utility maximizers.  
}%
EU often does not describe how people actually behave when they make
decisions~\cite{res}; thus EU is of limited utility if we want to model (and
perhaps predict) how people will behave.
As a result, 
many 
alternatives to \seu\ 
have been proposed in the literature
(see, for example,
\cite{Gul1991,GS1989,GiangShenoy01,KT1979,Luce00,Quiggin93,Schmeidler89,TK92,Yaari1987}).
Some of these rules involve representations of beliefs by means other
than a (single) probability measure;
\commentout{
(see, for example,
\cite{Gul1991,GS1989,GiangShenoy01,Quiggin93,Schmeidler89,Yaari1987});
}%
in some cases, beliefs and tastes are combined in ways other than the
standard way which produces expected utility;
yet other cases, such as 
\maximin\ 
and 
\commentout{
the rule of minimizing
regret~\cite{res}, 
}%
\reg~\cite{res},
do not require a representation of beliefs at all.

In
\cite{CH03a}, we propose a general framework in
which to 
study
and compare decision rules.  The idea is to define a
generalized notion of expected utility (\gseu), where a DM's
beliefs are \represented\ by 
\plmsrs\ \cite{FrH7} and the DM's tastes are
\represented\ by 
general (\ie\ not necessarily real-valued)
\utfns\@.  We show there that 
every \prefrel\ on acts has a \gseu\ \representation.
Here we show that \gseu\ is 
universal 
in a much stronger sense: we show that essentially all
decision rules 
have \gseu\ \representations.
\commentout{
This notion of one decision
rule being able to simulate another 
(much the same way that a Turing machine can simulate another)
seems to be novel.  
}%
The notion of \representing\ one \drule\ using another seems to be novel.
Intuitively, \drules\ are functions from tastes (and beliefs) to
\prefrels, so a \representation\ of a \drule\ is a \representation\ of a
\emph{function}, not a \prefrel. 

\commentout{
To formalize 
the notion of a \drule, 
we introduce the notion of a 
\dprob. Intuitively, 
a \emph{\dprob}
$\dprb$ describes all the relevant information facing a 
\dmaker\@: \ie\ a 
set of \acts\ available, a set of \ssofn, 
a set of \consequences, a \utfn, and possibly a \plmsr,
depending on the
\drule\ being used (\eg\  \maximin\ does not require a \plmsr\ while
\seu\ does).
\ijcai{
\footnote{Since \plmsrs\
generalize many different 
\representations\ of uncertainty, we consider only \drules\ that \represent\
uncertainty
using plausibility measures.
All the \drules\ mentioned 
earlier that use a representation of the DM's beliefs can be viewed
as using plausibility measures. For example,
\Mmeu\ of Gilboa and Schmeidler~\citeyear{GS1989}  \represents\
uncertainty using a set of probability measures, 
which 
can be viewed as a \plmsr.}
}%
A \drule\ can then be viewed as a function from \dprobs\ 
to \prefrels\ on the \acts.  That is, given the tastes and beliefs of
the DM, a decision rule returns a preference relation on acts.
}%

Roughly speaking, 
\commentout{
a \drule\ $\dr_1$ 
\emph{\represents}
a \drule\ $\dr_2$ 
iff a 
user of $\dr_2$
can be viewed as a user of $\dr_1$.
}%
given two \drules\ $\dr_1$ and $\dr_2$, an 
\emph{$\dr_1$ \representation\ of $\dr_2$} is 
a function $\tau$ that maps inputs of $\dr_2$ to inputs of $\dr_1$ that
contain the same \representation\ of tastes (and beliefs) such that $\dr_1(\tau(x)) = \dr_2(x)$.
Thus, $\trans$ models, in a precise sense, a user of $\dr_2$ as a user of
$\dr_1$, since $\trans$ preserves tastes (and beliefs).
We show 
that 
\commentout{
\gseu\ can \represent\ a large 
collection
of 
\drules,
}%
a large collection of \drules\ have \gseu\ \representations\
and 
\commentout{
provide a characterization of the \drules\
that it can 
\represent.
}%
characterize the collection.
\commentout{
A \drule\ turns out to be \representable\ by \gseu\ iff it is
}%
Essentially, a \drule\ has a \gseu\ \representation\ iff it is 
\emph{\uniform} 
in a precise sense.
It turns out that there are well-known \drules, such as maximizing
Choquet expected utility (CEU) \cite{Schmeidler89} 
that 
have no \gseu\ \representations.%
\footnote{The CEU decision rule is the appropriate one to use if belief
is represented by a Dempster-Shafer belief function; see
Section~\ref{sec:drules} for more discussion.}
This is because $\trans$ is not allowed to modify the 
representation of the tastes (and beliefs). 
We then define a notion of \emph{ordinal representation}, in which
$\trans$ is allowed to modify the representation of the tastes (and beliefs),
and is required to 
preserve only the ordinal aspect of the tastes (and beliefs).
We show that almost all \drules, including CEU, have 
ordinal
\gseu\
\representations. 
\commentout{
define what it means for 
We then relax the requirement that the output of $\trans$ must contain the same
\emph{representation} of the tastes (and beliefs) and allow $\trans$ to change
the representation while leaving the tastes (and beliefs) the same.
}%
\commentout{
We introduce a somewhat more general notion
of simulation called \emph{emulation}, which considers only 
a DM's 
\emph{ordinal} tastes and preferences (that is, it
considers, for example, only the preference 
relation
on consequences, and not
how much one consequence is preferred to another).  We 
show that essentially all 
decision rules (including CEU) can be emulated by 
\gseu\@.
}%

It is important to distinguish the result of \cite{CH03a}, which shows
that every {\em  \prefrel\/} can be represented by \gseu, from the
results of this paper, which show that many {\em \drules\/} can be
represented by \gseu, and almost all \drules\ can be ordinally
represented by \gseu.   Representing 
a \prefrel\ is not the same as representing a \drule.  Recall that a
\drule\ is a function from tastes (and possibly beliefs) to
preference relations on alternatives. \Drule\  $\dr$ represents a
\prefrel\ $\preceq$
if there are some tastes and beliefs such that, with these as input,
$\dr$ returns $\preceq$.  On the other hand, $\dr_1$ represents
$\dr_2$ if, roughly speaking, for {\em all\/} possible inputs of
tastes (and beliefs), $\dr_1$ and $\dr_2$ return the samqe
$\prefrel$. That is, $\dr_1$ and $\dr_2$ act essentially the same
way as functions.

There seems to 
be
no prior work in the literature that considers
how one \drule\ can \represent\ another.
Perhaps the
closest results to our own are those of Lehmann \citeyear{Lehmann2001}.
He proposes a ``unified general theory of
decision'' 
that contains both quantitative and qualitative decision theories.  
He considers a particular  \drule\ he calls \emph{\equm}, which allows
utilities 
to be nonstandard real numbers; he 
\ijcai{(\ala\ Robinson~\citeyear{Robinson1966}).  
More precisely, he allows the range of the utility function to be any
model (standard or nonstandard) of the real numbers,}%
defines a
certain 
preorder
on the nonstandard reals and makes decisions
based on 
maximizing 
expected utility (with respect to that 
preorder).
That his framework 
has \seu\ as a special case is immediate, since for the standard reals,
his 
preorder
reduces to the standard order on the reals.  He argues
informally that 
\maximin\ is a special case of his approach,
so that his approach can capture aspects of more qualitative decision
making as well.
\ijcai{
\footnote{Although Lehmann~\citeyear{Lehmann2001} claims to show how his
approach captures Maximin, in fact, his discussion shows how it captures
Maximax, not Maximin.  Nevertheless, his framework can capture Maximin
in a similar way [Lehmann, private communication, 2001].}
}%
It is easy to see that Lehmann's approach is a special case of \gseu;
his rule is clearly not universal in our sense.

The rest of this paper is organized as follows.  We cover some basic
definitions in Section~\ref{sec:preliminary}: expectation domains,
decision problems, \gseu, and decision rules (some of this material is
taken from \cite{CH03a}).  We show that \gseu\ can simulate any
decision rule in Section~\ref{sec:drule}, using Savage's act framework.
In Section~\ref{sec:lottery}, we show how these results can be applied
to the lottery framework originally introduced by von Neumann and
Morgenstern \citeyear{vNM1947} and Anscombe and Aumann's
\citeyear{AA1963} {\em horse lotteries}.  We conclude in
Section~\ref{sec:discussion} with some discussion of these results.
Proofs are deferred to the appendix.

\section{Preliminaries}\label{sec:preliminary}
To make this paper self-contained, much of the material in the first
three subsections of this section is taken 
(almost verbatim) from \cite{CH03a}.

\subsection{Plausibility, Utility, and Expectation Domains}

Since one of the goals of this paper is to provide a general framework for
all of decision theory, we want to \represent\ the \tandb\ of the \dmakers\ in
as general a framework as possible.  To this end, 
we use \plmsrs\ to
\represent\ the beliefs of the \dmakers\ and (generalized) \utfns\ to
\represent\ their tastes.

A \emph{\plstr} is a set $\pdom$, 
partially ordered by 
$\pleq_{\pdom}$  
(so $\pleq_{\pdom}$ is a reflexive, antisymmetric, and transitive
relation), with two 
special elements $\bot_{\pdom}$ and 
$\top_{\pdom}$, such that
(We often omit the subscript $\pdom$ in $\bot_\pdom$ and $\top_\pdom$
when it is clear from context.)
\commentout{
\mbox{for all $x \in \pdom$},
${\bottom_\pdom} \pleq_\pdom x \pleq_\pdom {\top_\pdom}$.
}%
${\bottom_\pdom} \pleq_\pdom x \pleq_\pdom {\top_\pdom}$ 
for all $x \in \pdom$.
A function $\plf : 2^\Stt \rar \pldom$ is a \emph{\plmsr} iff
\ben
\renewcommand{\theenumi}{$\plf$\arabic{enumi}}
\settowidth{\itemindent}{\theenumi}
\bl
$\plf(\eset) = {\bottom}$, 
\bl
$\plf(\Stt) = {\top}$, and 
\bl
if $X \sbset Y$ then $\plf(X) \pleq \plf(Y)$.
\een
As pointed out in~\cite{FrH7}, plausibility measures generalize, not
only  probability, but a host
of other \representations\ of uncertainty as well.  
A \emph{\utstr} is a set $\udom$ endowed with a 
reflexive binary relation
$\eleq_{\udom}$.
Intuitively, elements of $\udom$ \represent\ the strength of likes
and dislikes of the \dmaker\ while elements of $\pdom$ \represent\ the
strength of her beliefs.

Once we have plausibility and utility, 
we want to combine them to form expected utility.  
To do this, we introduce \expstrs, which have utility domains,
 plausibility domains, and operators
$\ep$ 
(the analogue of $+$)
and $\et$ 
(the analogue of $\times$).\footnote{We sometimes use $\times$ to denote
Cartesian product; the context will always make it clear whether this is the case.} 
More formally, 
an \emph{\expstr} 
is a 
tuple $\edom = (\udom, \pldom, \eudom, \et, \ep)$, where 
$(\udom, \eleq_\udom)$ is a \utstr, $(\pldom,\pleq_\pldom)$ is a \plstr,
$(\eudom, \eleq_\eudom)$ is a \valstr\ (where $\eleq_\eudom$ is a
reflexive binary relation),  
$\et : \pldom \times \udom \rar \eudom$, and 
$\ep : \eudom \times \eudom \rar \eudom$.
We have four requirements on \expstrs:
\ben 
\renewcommand{\theenumi}{$\edom$\arabic{enumi}}
\settowidth{\itemindent}{\theenumi}
\setlength{\itemsep}{0in}
\bl 
\label{edom-eq-first}
\label{edom-eq1}
$(x \ep y) \ep z  =  x \ep (y \ep z)$;
\bl 
\label{edom-eq2}
$x \ep y  =  y \ep x$;
\bl
\label{edom-eq3}
${\top} \et x = x$;
\bl
\label{edom-eq-last}
\label{edom-eq4}
$(\udom, \eleq_{\udom})$ is a substructure of $(\eudom, \eleq_\eudom)$.
\een
\ref{edom-eq1} and \ref{edom-eq2} say that $\ep$ is associative and
commutative.  
\ref{edom-eq3} says that $\top$  is the left-identity of $\et$ and
\ref{edom-eq4} ensures that the \expstr\ respects the relation on
utility values.

\ijcai{Note that we do not require that $\ep$  be monotonic; that is, 
we do not require that
if $x \eleq_\eudom y$ then $x \ep z \eleq_\eudom y \ep z$.  It turns out that
monotonicity does not really make a difference;
see comments after \mthm{thm:anyorder}.
}%

The \emph{standard expectation domain},
which we denote $\sedom$,
is 
$(\IR,[0,1],\IR,+,\times)$,
where the ordering on each domain is the standard order on the reals.

\subsection{Decision Situations and Decision Problems}

A \emph{\dsitu} describes the objective part of the circumstance
that the 
\dmaker\ faces (\ie\ the part that is independent of the tastes and beliefs of
the \dmaker)\@.
Formally, a \dsitu\ is
a tuple $\ACT = (\Act, \Stt, \Csq)$, where 
\commentout{
\bit
\bl $\Stt$ is the set of \ssofn, and
\bl $\Csq$ is the set of \consequences.
\bl $\Act$ is a set of acts (\ie\ 
a set of 
functions from $\Stt$ to $\Csq$),
\eit
}%
\bit
\bl $\Stt$ is the set of \ssofn, 
\bl $\Csq$ is the set of \consequences, and
\bl $\Act$ is a set of acts (\ie\ 
a set of 
functions from $\Stt$ to $\Csq$).
\eit
An act $\act$ is \emph{simple} iff 
its range is finite.
That is, $\act$ is simple if it has only finitely many \consequences.
Many works in the literature focus on simple \acts\
(\eg\ \cite{Fishburn1987}).
We assume in this paper that $\Act$ contains only simple acts; this means
that we can define (generalized) expectation using 
finite sums, so we do not have to introduce infinite series or integration for
arbitrary \expstrs.
Note that all  \acts\ are guaranteed to be simple if either $\Stt$ or
$\Csq$ is finite, although we do not assume that here.

A \dprob\ is essentially a \dsitu\ together with information about the 
tastes (and beliefs)
of the \dmaker;  that is, a \dprob\ is a \dsitu\ together with the subjective
part of the circumstance that faces the \dmaker\@.
Formally, a 
\emph{\nonplstic\ \dprob} 
is a tuple 
$(\ACT, \udom, \utf)$, where
\bit
\bl $\ACT = (\Act, \Stt, \Csq)$ is a decision situation,
\bl $\udom$ is a utility domain, and
\bl $\utf : \Csq \rar \udom$ is a \utfn.
\eit
A \emph{\plstic\ \dprob} is a tuple
$(\ACT, \edom, \utf, \plf)$, where
\bit
\bl $\ACT = (\Act, \Stt, \Csq)$ is a decision situation,
\bl $\edom = (\udom,\pldom,\eudom,\et,\ep)$ is an \expstr, 
\bl $\utf : \Csq \rar \udom$ is a \utfn, and
\bl $\plf : 2^\Stt \rar \pdom$ is a \plmsr.
\eit
We could have let a \plstic\ \dprob\ be simply a
\nonplstic\ \dprob\ together with a \plstr\ and a \plmsr, without
including the 
other components of expectation domains.
However, this turns out to complicate the 
presentation (see below).

We say that $\dprb$ is \emph{standard} iff
its \utstr\ is $\realNum$ (and, if $\dprb$ is \plstic, 
its \plmsr\ is a probability measure and its \expstr\ is $\sedom$).

\subsection{Expected Utility}
\label{sec:expectedutility:gseu}

Let $\dprb$ be a \dprob\ with $\Stt$ as the set of states, $\udom$ as the
\utstr, and $\utf$ as the \utfn. 
Each act $\act$ of $\dprb$ induces a
\emph{\urv}
$\utf_{\act} : \Stt \rar \udom$ 
as follows:
$\utf_{\act}(s) \bydef{=} \utf(\act(s))$.
If in addition $\dprb$ is \plstic\ with $\pdom$ as the \plstr\ and 
$\plf$ as the \plmsr, 
then each $\act$ also induces a \emph{\ulottery}
$\lto^{\plf,\utf}_\act : \ran(\utf_{\act}) \rar \pdom$ as follows:
$\lto^{\plf,\utf}_\act(u) = \plf(\utf^{-1}_\act(u))$.
Intuitively, $\lto^{\plf,\utf}_\act(u)$ is the likelihood of getting
utility $u$ when 
performing act $\act$.
If $\dprb$ is in fact standard  (so 
$\edom = \sedom$
and $\plf$
is a probability measure $\Pr$), we can identify the 
expected utility of act $\act$ with the expected value of
$\utf_{\act}$
with respect to $\Pr$,
computed in the 
standard
way:
\beqn
\label{eqn:std-meu}
\eprut{\act} \bydef{=} 
\sum_{x \in \ran(\utf_{\act})} \Pr(\utf_{\act}^{-1}(x)) 
\times x.%
\ijcai{\footnote{If the domain of $\Pr$ is some
we must assume that $\utf_{\act}$ is a measurable function; that is,
$\utf_{\act}^{-1}(x)$ is a measurable set for all $x \in \ran(\utf_{\act})$.}}%
\eeqn
\commentout{
This expression for expected utility has an obvious generalization in an
arbitrary \expstr\ $\edom = (\udom, \pldom, \eudom, \et, \ep)$, 
}%
As we mentioned earlier, since acts are assumed to be simple, this sum
is finite.
We can generalize \meqn{eqn:std-meu} to an arbitrary \expstr\ 
$\edom = (\udom, \pldom, \eudom, \et, \ep)$ by
\commentout{
where $+$ is 
replaced by $\ep$, 
$\times$ is replaced by $\et$, 
and $\Pr$ is
replaced by $\plf$.  
}%
replacing $+$, $\times$, and $\Pr$ by
$\ep$, $\et$, and $\plf$, respectively.
This gives us
\beqn
\label{eqn:def-E}
\eplut{\act} \bydef{=}
\EP_{x \in \ran(\utf_{\act})}
\plf(\utf_{\act}^{-1}(x)) \et x.
\eeqn
We call (\ref{eqn:def-E}) the
\emph{generalized \seu} (\gseu)
of act $\act$.
Clearly \meqn{eqn:std-meu} is a special case of \meqn{eqn:def-E}.

\commentout{
Given a \plstic\ $\dprb$, each $\act$ of $\dprb$ also induces a \emph{\ulottery}
$\lto^{\plf,\utf}_\act : \ran(\utf_{\act}) \rar \pdom$ as follows:
$\lto^{\plf,\utf}_\act(u) = \plf(\utf^{-1}_\act(u))$. 

Note that
$\eplut{\act}$ is just the expected value of $\lto^{\plf,\utf}_\act$.
}%

\subsection{\DRules}\label{sec:drules}

Intuitively, a \drule\ tells the \dmaker\ what to do 
when facing a \dprob\ in order to get a \prefrel\ on acts---\eg\ compare 
the expected utility of \acts.
Just as we have \nonplstic\ \dprobs\ and \plstic\ \dprobs, we have
\nonplstic\ \drules\ and \plstic\ \drules. 
As the name suggests, (non)\plstic\ \drules\ are defined
on (non)\plstic\ \dprobs.  

We do not require decision rules to be defined on all decision
problems.  
For example, (standard) \seu\ is defined only on 
\commentout{
plausibilistic decision
problems where 
the expectation domain is 
$\sedom$
and the plausibility measure is a
probability measure.
(We call these \emph{probabilistic decision problems}.)
}%
standard \plstic\ \dprobs.
More formally, a 
\emph{(non)\plstic\ \drule} 
$\dr$
is a 
\commentout{
partial function from 
the collection of (non)\plstic\ \dprobs\ to \prefrels\ on
their \acts.
Let $\dom(\dr) = \{ \dprb \st \dr(\dprb) \neq \ndef \}$.
}%
function whose domain, denoted $\dom(\dr)$, is a 
set of (non)\plstic\
\dprobs, and whose range, denoted $\ran(\dr)$, is 
a
set of \prefrels\ on acts.
If $\dprb \in \dom(\dr)$
and $\act_1$ and $\act_2$ are acts
in $\dprb$, then 
we write 
\commentout{
\begin{center}
$\act_1 \eleq_{\dr(\dprb)} \act_2$ 
iff
$(\act_1, \act_2) \in \dr(\dprb)$.
\end{center}
}%
\[
\act_1 \eleq_{\dr(\dprb)} \act_2
\mbox{ iff }
(\act_1, \act_2) \in \dr(\dprb).
\]
\ijcai{
\footnote{Readers 
familiar with set theory will note that the collection of all \dprobs\
(\plstic\ or \nonplstic) is not a set, but a proper class.  We can get around
this problem by relativizing to sets, but this would complicate the
presentation.  For ease of exposition, we ignore the issue of proper
classes in this paper.}
That is, 
taking $\ndef$ to denote ``undefined'', 
if $\dr$ is a 
(non)\plstic\ \drule\ and 
$\dprb$ is a 
(non)\plstic\ \dprob, then 
$\dr(\dprb)$ is relation on the acts of $\dprb$ if $\dr$ is defined
on $\dprb$ and $\dr(\dprb) = \ndef$ if $\dr$ is not defined on $\dprb$ (\eg\ if
$\dr = \mr{\maximin}$ and 
the \utstr\ of $\dprb$
is not linearly 
ordered).  
}%
Here are a few  examples of decision rules:
\commentout{
\begin{list}{$\bullet$}{
\setlength{\itemindent}{0.25in}
\setlength{\leftmargin}{0in}}
}%
\bit
\item
\gseu\ is a \plstic\ decision rule whose domain 
consists of all
plausibilistic decision problems.  Given a 
\plstic\ 
decision problem 
$\dprb = (\ACT, \edom, \utf,\plf)$, where
$\edom = (\udom,\pdom,\eudom,\ep,\et)$,
we have 
\commentout{
$\act_1 \eleq_{\mr{\gseu}(\dprb)} \act_2$ iff 
$\eplut{\act_1} \eleq_{\eudom} \eplut{\act_2}$ 
}%
\[
\act_1 \eleq_{\mr{\gseu}(\dprb)} \act_2 \mbox{ iff }
\eplut{\act_1} \eleq_{\eudom} \eplut{\act_2}
\]
for all acts
$\act_1, \act_2$ 
in
$\ACT$.  Note that \gseu\ would not be a \drule\
according to this 
definition if \plstic\ \dprobs\ contained only a utility function and a
plausibility measure, and did not include the 
other components of 
expectation domains.  
\item
Of course, standard \seu\ is a 
\drule\ (whose domain consists of all 
standard \plstic\ \dprobs).
\item \maximin\ is a nonplausibilistic decision rule that orders 
acts
according to their worst-case consequence.  
It is a conservative rule; 
the ``best'' act according to \maximin\ is the one with the best worst-case
consequence.  
Intuitively, \maximin\ views Nature as an adversary that always pick a
state that realizes the worst-case consequence, no matter what act the
\dmaker\ chooses.
The domain of (standard) \maximin\
consists of nonplausibilistic decision problems with real-valued
utilities.  Given an act $\act$ and a real-valued utility function
$\utf$, let 
$\worst_{\utf}(\act) = \min_{s \in \Stt}\utf_\act(s)$.
Then given a decision problem 
$\dprb = (\ACT, \realNum, \utf)$,
\commentout{
}%
\[
\act_1 \eleq_{\mr{\maximin}(\dprb)} \act_2 \mbox{ iff }
\worst_{\utf}(\act_1) \leq \worst_{\utf}(\act_2).
\]
Clearly the domain of \maximin\ can be extended so that it includes all
nonplausibilistic decision problems where the range of the utility
function is totally ordered.

\item 
\reg\ (REG)
is based on a different philosophy.  It tries
to hedge a DM's bets, by doing reasonably well no matter what the
actual state is.   It is also a
nonplausibilistic rule. As a first step to defining it, given a
nonplausibilistic decision 
problem 
$\dprb = ((\Act,\Stt,\Csq), \realNum, \utf)$, 
\commentout{
$\sfa_{\stt}$ be an act in $\Act$ that
gives the best consequence in $\stt$; that is, 
$\utf_{\sfa_s}(s) \ge \utf_\sfa(s)$ for all acts $\sfa \in \Act$.  
}%
let $\best : \Stt \rar \udom$ be defined as
$\best(\stt) = \sup_{\act \in \Act} \utf_{\act}(\stt)$; that is,
$\best(\stt)$ is the least upper bound of the utilities in state $\stt$. 
The \emph{regret} of $\sfa$ in
state $s$, denoted 
$\regret(\sfa,s)$, is
$\best(\stt) - \utf_{\act}(\stt)$;
\commentout{
that 
is, the regret of $\sfa$ in $\stt$ is the difference between the utility of
performing the best act in
$\stt$ (the act that the DM would perform, presumably, if she knew
that the actual state was $s$) and that of performing $\sfa$ in $\stt$.
}%
note that no act can do better than $\act$ by more than
$\regf(\act,\stt)$ in state $\stt$. 
Let 
$\overline{\regret}(\sfa) = \sup_{\stt \in \Stt} \regret(\sfa,\stt)$.
\commentout{
For example, if $\regret_\utf(\sfa) = 2$, then in each state $\stt$,
the utility of performing $\sfa$ in $\stt$ is guaranteed to be
within $2$ of the utility generated by the act the DM would choose if
she knew that the actual state was $\stt$.
}%
For example, suppose that $\overline{\regf}(\act) = 2$ and the DM picks $\act$.
Suppose that the DM then learns that the true state is 
$\stt_0$
and is offered 
a chance 
to change her mind.
No matter what 
act 
she picks, the utility of the new act cannot be more than 2
higher then $\utf_\act(\stt_0)$. 
REG
orders acts by their regret and thus takes the ``best'' act to be the
one that minimizes 
$\overline{\regf}(\act)$.
Intuitively, this rule 
tries
to minimize the regret that a DM would feel if she discovered what the
situation actually was: the ``I wish I had done 
$\sfa_2$ instead of $\sfa_1$'' feeling. 
Thus, 
\commentout{
$\act_1 \eleq_{\mr{REG}(\dprb)} \act_2$ iff
$\overline{\regret}(\sfa_1) \ge \overline{\regret}(\sfa_2)$.  
}%
\[
\act_1 \eleq_{\mr{REG}(\dprb)} \act_2 \mbox{ iff }
\overline{\regret}(\sfa_1) \ge \overline{\regret}(\sfa_2).  
\]
Like \maximin, Nature is viewed as an adversary that would pick a state that
maximizes regret, no matter what act the \dmaker\ chooses.
It is well known that, 
in general, 
\maximin, REG,
and \seu\ give different
recommendations~\cite{res}.  

\item The Maxmin Expected Utility rule (MMEU) \cite{GS1989} assumes
that a DM's beliefs are represented by a set $\P$ of probability
measures.  Act $\act_1$ is preferred to $\act_2$ if the worst-case
expected utility of $\act_1$ (taken over all the probability measures in
$\P$) is at least as large as the worst-case 
expected
utility of $\act_2$. 
Thus MMEU is, in a sense, a hybrid of \seu\ and \maximin.
To view MMEU as a function on decision problems, we must first show how
to represent a set of probability measures as a single plausibility
measure.  
We do this using an approach due to Halpern \citeyear{Hal25}. 
Let the plausibility domain $\pdom = [0,1]^{\P}$, that is, all functions
from $\P$ to $[0,1]$, ordered pointwise; 
in other words,
$p \pleq_\pdom q$ iff
$p(\Pr) \leq q(\Pr)$ for all $\Pr \in \P$.  Thus, in this domain, $\bot$
is the constant function 0 and $\top$ is the constant function 1.
For each $X \sbset \Stt$,
let $f_X \in \pdom$ be the function that evaluates 
each probability measure in $\P$ at $X$; that is, $f_X(\Pr) = \Pr(X)$ for all
$\Pr \in \P$.  
Let $\Pl_{\P}(X) = f_X$; it is easy to verify that $\plf_\P$ is a \plmsr.
We view $\Pl_{\P}$ as a representation of 
the set $\P$ of probability measures;
clearly $\P$ can be recovered from $\Pl_{\P}$.
The domain of MMEU consists of all plausibilistic decision problems of
the form 
$\dprb = ((\Act,\Stt,\Csq),(\realNum, [0,1]^{\P}, \eudom, \ep, \et),\utf,\Pl_{\P})$,
where $\P$ is a set of probability measures
on $2^\Stt$, and
\commentout{
$\act_1 \eleq_{\mr{\mmeu}(\dprb)} \act_2$ 
iff
$\inf_{\Pr \in \P} \eprut{\act_1} \le \inf_{\Pr \in \P} \eprut{\act_2}$.  
}%
\[
\ts
\act_1 \eleq_{\mr{\mmeu}(\dprb)} \act_2
\mbox{ iff }
\inf_{\Pr \in \P} \eprut{\act_1} \le \inf_{\Pr \in \P} \eprut{\act_2}.  \
\]
Note that this definition ignores
$\ep$, $\et$, and $\eudom$.

\item 
A \emph{nonadditive probability} \cite{Schmeidler89} $\nu$ is just a function
associating with each subset of a set $\Stt$ a number between 0 and 1,
where $\nu(\emptyset) = 0$, $\nu(\Stt) = 1$, and $\nu(X) \le \nu(Y)$ if
$X \subseteq Y$.    (Roughly speaking, a nonadditive probability is just
a plausibility measure whose range is $[0,1]$, where $\bot = 0$ and
$\top = 1$.)
Schmeidler \citeyear{Schmeidler89} used a notion of expected utility for
nonadditive probability that was defined by Choquet \citeyear{Choq}.
(Choquet applied his notion of expectation to what he called {\em
capacities}; nonadditive probabilities generalize capacities.)  
Given
an act $\act$, a real-valued utility function $\utf$
such that
$\ran(\utf_\act) = \{ u_1, \ldots, u_n \}$ and 
$u_1 < \cdots < u_n$, 
and a nonadditive probability $\nu$, define
\begin{equation}\label{eq:ebel}
\E_{\nu}(\utf_{\act}) = u_1 + \sum_{i=2}^n 
\nu(X_i) \times (u_i - u_{i-1}),
\end{equation}
where $X_i = \utf_{\act}^{-1}(\{u_i, \ldots, u_n\})$.
It is easy to check (\ref{eq:ebel}) agrees with (\ref{eqn:std-meu}) if 
$\nu$ is a probability measure.  
The Choquet expected utility (\ceu) rule has as its domain decision
problems of the form  
$\dprb = (\ACT,\sedom, \utf, \nu)$, and it orders acts as follows:
\[
\act_1 \eleq_{\mr{\ceu}(\dprb)} \act_2 \mbox{ iff }
\E_{\nu}(\utf_{\act_1}) \le  \E_{\nu}(\utf_{\act_2}).  
\]

A special case of a nonadditive probability is a Dempster-Shafer \emph{belief
function} \cite{Demp1}.  Belief functions also generalize probability.
That is,
every probability measure is a belief function, but the converse is not
necessarily true.%
\footnote{We assume that the reader is familiar with belief functions;
see \cite{Shaf} for details.  In any case, a knowledge of belief
functions is not necessary for understanding the results of this paper.}
Given a belief function
$\Bel$, it is well-known that there exists a set $\P_{\Bel}$ of
probability measures such that for all $X \subseteq \Stt$, 
$\Bel(X) = \inf_{\Pr \in \P_{\Bel}} \Pr(X)$ \cite{Demp1}.  
Moreover, if we use the \ceu\ rule to compute expected belief, then it
follows from results of Schmeidler \citeyear{Schmeidler86} that
\begin{equation}\label{ebel=le}
\E_{\Bel}(\utf_{\act}) = 
\ts \inf_{\Pr \in \P_\Bel} \eprut{\act}.
\end{equation}

Let $\dprb = (\ACT,\sedom, \utf, \Bel)$ and let $\dprb_{\P_\Bel}$ 
be the decision problem that results from $\dprb$ by replacing
$\Bel$ by $\plf_{\P_{\Bel}}$, 
It is immediate from
(\ref{ebel=le}) that if 
$\dprb_{\P_\Bel}$ 
is the decision problem that results from $\dprb$ by replacing
$\Bel$ by 
$\plf_{\P_{\Bel}}$, 
and replacing the plausibility domain $[0,1]$ 
in the expectation  domain by 
$[0,1]^{\P_\Bel}$, 
then 
$\act_1 \eleq_{\mr{\ceu}(\dprb)} \act_2$ iff 
$\act_1 \eleq_{\mr{\mmeu}(\dprb_{\P_\Bel})} \act_2$.%
\footnote{It follows from results of Schmeidler \citeyear{Schmeidler86}
that a similar result holds, not just for belief functions, but for a
larger set of nonadditive probability measures.  Say that a probability
measure $\Pr$ dominates a nonadditive probability $\nu$ on $\Stt$ if
$\Pr(X) \ge \nu(X)$ for all $X \subseteq \Stt$.  The result holds for
all $\nu$ such that $\nu = \inf\{\Pr: \Pr$ dominates $\nu\}$.}
\eit

\section{\Representing\ \DRules}
\label{sec:drule}

Given a \drule\ $\dr$ and a \prefrel\ $\leqact$ on the set of acts $\Act$,
an \emph{$\dr$ \representation\ of $\leqact$} is basically a \dprob\
$\dprb \in \dom(\dr)$ such that
$\dr(\dprb) = {\leqact}$ (and the set of acts in $\dprb$ is $\Act$).
In other words, an $\dr$ \representation\ of $\leqact$ makes
$\dr$ relate acts in $\Act$ the way $\leqact$ relates them, so 
we can model a \dmaker\ whose \prefrel\ is $\leqact$ as a user of $\dr$. 
In \cite{CH03a} 
we prove the following:
\thm\label{thm:anyorder} 
Every preference relation
$\leqact$ has a \gseu\ \representation.
\ethm
We then go on to show how constraints on \gseu\ can be used to capture
various postulates on preference relations, such as Savage's postulates
\cite{Savage}. 

In this paper, we go in a somewhat different direction.
We start by extending the notion of \representation\ to \drules.
Intuitively, we want an $\dr_1$ \representation\ of $\dr_2$ to allow us
to model a user of 
$\dr_2$ as a user of $\dr_1$. 
We then investigate the extent to which \gseu\ can represent arbitrary
\drules. 
To make this precise, we need a few definitions. 

\commentout{
We show here that (almost) all \drules\ have \gseu\ \representations.  

Roughly speaking, 
a \drule\ $\dr_1$ \represents\ another \drule\
$\dr_2$ iff \mbox{$\dr_1(\dprb) = \dr_2(\dprb)$} for all 
$\dprb \in \dom(\dr_2)$.
However, this is somewhat too 
restrictive, 
since some \drules\ are \nonplstic\ while others are \plstic; we
want to allow \plstic\ \drules\ to \represent\ \nonplstic\ 
\drules.
To make this precise, we need a few definitions.
given two \drules, a $\dr_1$ \representation\ of $\dr_2$ 
}%
Two (\plstic) \dprobs\ $\dprb_1$ and $\dprb_2$  
are 
\emph{\congruent}, denoted $\dprb_1 \cong \dprb_2$, 
iff they involve the same \dsitu, 
\commentout{
\utfn, and (if both are \plstic) the same \plmsr. 
}%
\utstr, 
and \utfn\
(and, if both are \plstic, the same 
\plstr\ and \plmsr\ as well). 
Note that if $\dprb_1 \cong \dprb_2$,
then they
agree on the tastes 
(and beliefs) of the \dmaker,
\commentout{
In particular, 
if $\dprb_1$ and $\dprb_2$ are \nonplstic, then
$\dprb_1 \cong \dprb_2$ iff $\dprb_1 = \dprb_2$.
the way expressions are evaluated and related.
}%
so if they are both \nonplstic, then $\dprb_1 = \dprb_2$, and if they are both
\plstic, then they differ only in 
the $\eudom$, $\eleq_\eudom$, $\ep$, and $\et$ components of their 
\expstrs.

A \emph{\druletr} $\trans$ is a function that maps
inputs of one \drule\ $\dr_2$ to the inputs of another rule $\dr_1$.
A \druletr\ $\trans$ is an \emph{$\dr_1$ \representation\ of $\dr_2$} iff
$\dom(\trans) = \dom(\dr_2)$ and 
for all $\dprb \in \dom(\dr_2)$,
\commentout{
\begin{center}
$\trans(\dprb) \cong \dprb$ and 
$\dr_1(\trans(\dprb)) = \dr_2(\dprb)$.
\end{center}
}%
\commentout{
\[
\trans(\dprb) \cong \dprb \mbox{  and }
\dr_1(\trans(\dprb)) = \dr_2(\dprb).
\]
}%
\bit
\bl $\trans(\dprb) \cong \dprb$ and 
\bl $\dr_1(\trans(\dprb)) = \dr_2(\dprb)$.
\eit
\commentout{
\Drule\ $\dr_1$ \emph{\represents} $\dr_2$
iff there exists a \dprob\ transformation $\trans$ such that
$\dr_1(\trans(\dprb)) = \dr_2(\dprb)$ 
for all $\dprb \in \dom(\dr_2)$.
}%
\commentout{
Thus a \dmnuser of $\dr_2$ 
as if she is using $\dr_1$, 
}%
Thus a \dmaker\ that uses $\dr_2$ to relate acts based on her tastes (and
beliefs) behaves as if she is using $\dr_1$, 
since $\trans(\dprb) \cong \dprb$ and $\dr_1(\trans(\dprb)) = \dr_2(\dprb)$. 
\ijcai{
The reason that we want $\trans(\dprb)$ to be \congruent\ with $\dprb$
is that we do not want to tamper with the DM's tastes and beliefs.}%

\commentout{
Clearly \gseu\ can represent 
\seu\@.  
We now show it can represent some
other rules of interest.
}%
Note that $\trans(\dprb) = \dprb$ is a \gseu\ \representation\ of \seu\@. 
We now consider some less trivial examples.

\xam
To see that \maximin\ has a \gseu\ \representation, 
let 
$\edom_{\mr{max}} = (\realNum, \{0,1\}, \realNum \union \{\infty\},
\min, \et),$ let 
$\Pl_{\mr{max}}$ be the plausibility measure such that 
$\Pl_{\mr{max}}(X)$ is 0 if 
$X = \emptyset$ and 1 otherwise,
and define $1 \et x = x$ and $0 \et x = \infty$.
If $\dprb = (\ACT,\realNum,\utf),$ where $\ACT = (\Act,\Stt,\Csq)$, 
then  it is easy to check that 
$\E_{\Pl_{\mr{max}},\edom_{\mr{max}}}(\utf_{\act}) = \worst_{\utf}(\act)$. 
Take 
$\tau(\dprb) = (\ACT,\edom_{\mr{max}},\utf,\Pl_{\mr{max}})$.  
Clearly
$\trans(\dprb) \cong \dprb$: the 
\dsitu\ and \utfn\ have
not changed.
Moreover, it is immediate that 
$\mr{\gseu}(\tau(\dprb)) = \mr{\maximin}(\dprb)$. 
\ijcai{
(A similar argument shows that \gseu\ can represent a generalized
version of maximin, whose domain includes all decision problems whose
utilities are totally ordered.)}
\exam

\xam 
To see that \reg\ (REG) has a \gseu\ \representation, for ease of 
exposition, we take 
$\dom(\mr{REG})$
to consist of 
standard
decision problems 
$\dprb = ((\Act, \Stt, \Csq), \IR, \utf)$ 
such that 
$M_\dprb = \sup_{\stt \in \Stt} \best(\stt) < \infty$.
(If $M_{\dprb} = \infty$, given the restriction to simple acts, 
it
is 
easy to
show that 
all acts have infinite regret.)
Let 
$\edom_{\mr{reg}} = ( \IR, [0, 1], \IR \union \{\infty\}, \min, \et)$, where 
\commentout{
$x \et y = y - \log(x)$
if $x > 0,$ and $x \et y = 0$ if $x = 0$.
}%
\[
x \et y = 
\left\{\begin{array}{ll}
y - \log(x) & \mbox{if $x > 0$ and}\\
\infty & \mbox{if $x = 0$.}
\end{array}\right.
\]
Note that ${\bot} = 0$ and ${\top} = 1$.
Clearly, $\min$ is associative and commutative, and
${\top} \et r = r - \log(1) = r$ 
for all $r \in \IR$.  Thus,
$\edom_{{\rm reg}}$ is an expectation domain.

For $\emptyset \neq X \subseteq \Stt,$ define
$M_X = \sup_{s\in X}\best(\stt)$.
Note that 
$M_\Stt = M_{\dprb} < \infty$; also 
if $X \sbset Y$, then
$M_X \leq M_Y$.  
\commentout{
Thus
it  follows that $M_X < \infty$ for all $ X \subseteq \Stt$.   
Let $\Pl_\dprb$ be the plausibility measure such that 
$\Pl_\dprb(\emptyset) = 0$ and $\Pl_M(X) = e^{M_X- M_\dprb}$
for $X \neq \emptyset$.
}%
Let $\plf_\dprb(\eset) = 0$ and $\plf_\dprb(X) = e^{M_X - M_\Stt}$.
It is easy to verify that $\plf_\dprb$ is a \plmsr.
It is also easy to check that
\[
\E_{\Pl_\dprb,\edom_{\mr{reg}}}(\utf_{\sfa}) = M_{\dprb} - \overline{\regf}(\sfa)
\]
for all acts $\sfa \in \Act$.
Let
$\tau(\dprb) = (\ACT, \edom_{\rm{reg}}, \utf,\Pl_{\dprb})$.
\commentout{
Clearly, higher expected utility corresponds to lower regret, 
so 
$\mr{\gseu}(\tau(\dprb)) = \mr{REG}(\dprb)$;
furthermore, $\trans(\dprb) \cong \dprb$, since the \utfn\ has not changed.
}%
Clearly,
$\trans(\dprb) \cong \dprb$, 
since the 
\dsitu\ and \utfn\ have
not changed; 
furthermore,
$\mr{\gseu}(\tau(\dprb)) = \mr{REG}(\dprb)$,
since higher expected utility corresponds to lower regret.
\exam

\xam\label{xam:mmeu} 
To see that \mmeu\ has a \gseu\ \representation,
let 
$\dprb \in \dom(\mr{\mmeu})$ such that 
$\dprb = (\ACT,(\realNum, [0,1]^{\P}, \widehat{\eudom}, \widehat{\ep}, \widehat{\et}),\utf, \Pl_{\P})$. 
\commentout{ 
given a decision situation 
      $\ACT = (\Act, \Stt, \Csq)$, a utility 
function $\utf$, 
a standard \dprob\ $((\Act,\Stt,\Csq),\realNum,\utf)$
and a set 
$\P$ of probability measures, consider the expectation domain 
$\edom_\P = (\realNum, [0,1]^\P, \realNum^\P,\ep, \et)$, where, 
as before, $[0,1]^\P$ are ordered pointwise, and the elements of
$\realNum^\P$ are ordered according to their infs.  That is,
}%
Let 
$\edom_\P = (\realNum,[0,1]^\P, \realNum^\P, \ep,\et)$, where
\commentout{
$$\mbox{$f \eleq_\eudom g$ iff $\inf \{f(\Pr) \st \Pr \in \P\} \leq \inf
\{g(\Pr) 
\st \Pr \in \P\}$.}$$
Define $\ep$ as pointwise addition and $\et$ as scalar multiplication;
that is $(p \ep q)(\Pr) =
p(\Pr) + q(\Pr)$ for all $\Pr \in M$, and 
$(p \et u)(\Pr) = u \times p(\Pr)$ 
for all $\Pr \in M$.  
}%
$\ep$ is pointwise function addition, $\et$ is scalar multiplication, and
\[
\ts
f \eleq_{\realNum^\P} g \mbox{  iff  } \inf_{\Pr \in \P} f(\Pr)  \leq \inf_{\Pr \in \P}
g(\Pr).
\]
Note that we can
identify $\realNum$ with the constant functions in $\realNum^\P$, so
$\realNum$ can be viewed as a substructure of $\realNum^\P$.  With these
definitions, 
$\edom_\P$ 
is an expectation domain.  
\commentout{
Let $\dprb_\P = (\ACT,\edom_\P,\utf,\plf_\P)$.
It is immediate from the definition of $\eleq_\eudom$ 
that 
\[
\act \eleq_{\mr{\gseu}(\dprb_\P)} \actb \mbox{ iff } 
{\inf}_{\Pr \in \P} \eprut{\act} \leq {\inf}_{\Pr \in \P} \eprut{\actb}.
\]
Thus, \gseu\ represents 
\mmeu\@.
}%
Let $\trans(\dprb) = (\ACT,\edom_\P,\utf,\plf_\P)$.
It is immediate from the definition of 
$\eleq_{\realNum^\P}$ that
\[
\ts
\act \eleq_{\mr{\gseu}(\trans(\dprb))} \actb \mbox{ iff }
\inf_{\Pr \in \P} \eprut{\act} \leq \inf_{\Pr \in \P} \eprut{\actb}.
\]
Thus $\mr{\gseu}(\trans(\dprb)) = \mr{\mmeu}(\dprb)$; 
furthermore, it is clear that $\trans(\dprb) \cong \dprb$, since 
the \dsitu, \utfn, and \plmsr\ have
not changed.
\exam

Although it can represent many decision rules,
\gseu\ cannot represent 
\ceu\@.
We can in fact characterize the conditions under which a \drule\ is 
\representable\ by \gseu\@.

There is a trivial condition that a decision rule must satisfy in order
\commentout{
to be represented by 
\gseu\@.  A 
}%
for it to have a \gseu\ \representation. 
\commentout{
A
\drule\ $\dr$ respects utility if the preference 
relation
on acts
induced by $\dr$ agrees with the utility ordering.
}%
Intuitively, a \drule\ $\dr$ respects utility if
$\dr$ relates acts of constant utility according to the relation between
utility values. 
\commentout{
Formally, $\dr$ \emph{respects utility} 
iff for all 
$\dprb \in \dom(\dr)$
and 
$\act_i$ is $u_i$,  then 
$\act_1 \leqRdp{\dprb} \act_2$ iff  
$u_1 \eleq_\udom u_2$.
}%
Formally, 
a \drule\
$\dr$ \emph{respects utility} 
iff for all 
$\dprb \in \dom(\dr)$
with $\Act$ as the set of acts, $\Stt$ as the set of states, 
$\udom$ as the \utstr, and $\utf$ as the \utfn,
for all
$\act_1, \act_2 \in \Act$, if
$\utf_{\act_i}(\stt) = u_i$ for all states $\stt \in \Stt$,
then 
\beqn
\label{resputil}
\act_1 \leqRdp{\dprb} \act_2 \mbox{ iff }
u_1 \eleq_\udom u_2.
\eeqn
We say that $\dr$ \emph{weakly respects utility} iff \meqn{resputil} holds for
all constant acts (but not necessarily for all acts of constant utility).
\commentout{
It is easy to see that the  
\prefrels\ induced
by \gseu\ 
respects utility, since the expected utility of the constant
act with utility $u$ is just $u$.   (This depends on the assumption that
${\top} \et x = x$.)  Thus, \gseu\ cannot possibly
represent a decision rule that does not respect utility.
}%
It is easy to see that \gseu\ respects utility, 
since ${\top} \et u = u$ for all $u \in \udom$ and
$(\udom,\eleq_\udom)$ is a substructure of $(\eudom,\eleq_\eudom)$.
Thus if $\dr$ does not respect utility, it has no \gseu\ \representation. 
While respecting utility is a necessary condition for a decision rule to
have a \gseu\ \representation,
it is not sufficient.  It is also
necessary for the decision rule to treat acts that behave in similar
ways similarly.  
\commentout{
Recall that each $\act$ from a \dprob\ $\dprb$ induces a \urv\ $\utf_\act$,
where $\utf$ is the \utfn\ of $\dprb$.
If $\dprb$ is \plstic, $\act$ also induces a \emph{\ulottery} 
$\lto^{\plf,\utf}_\act : \ran(\utf_\act) \rar \pdom$ as follows:
$\lto^{\plf,\utf}_\act(u) = \plf(\utf^{-1}_\act(u))$, where 
$\pdom$ is the \plstr\ of $\dprb$ and $\plf$ is the
\plmsr\ of $\dprb$. 
}%
Two acts $\act_1, \act_2$ in a decision problem $\dprb$  are
\emph{indistinguishable},
denoted
$\act_1 \ind_{\dprb} \act_2$
iff either
\begin{itemize}
\item $\dprb$ is nonplausibilistic and
$\utf_{\act_1} = \utf_{\act_2}$, or
\item $\dprb$ is plausibilistic,
\commentout{
$\ran(\utf_{\act_1}) = \ran(\utf_{\act_2})$, and
$\Pl(\utf^{-1}_{\act_1}(u)) = \Pl(\utf^{-1}_{\act_2}(u))$ for
all utilities $u$ in the common range of $\utf_{\act_1}$ and $\utf_{\act_2}$,
where $\Pl$ is the plausibility measure in $\dprb$.
}%
and $\lto^{\plf,\utf}_{\act_1} = \lto^{\plf,\utf}_{\act_2}$,
\end{itemize}
where $\utf$ is the \utfn\ of $\dprb$ and $\plf$ is the \plmsr\ of $\dprb$. 
In the nonplausibilistic case, two acts are indistinguishable if they
induce the same 
\urv;
in the plausibilistic case,
they are indistinguishable if they induce the same 
\commentout{
plausibilistic
utility distribution.
}%
\ulottery.

A decision rule 
$\dr$
is 
uniform if it respects
indistinguishability.  
More formally, 
$\dr$
is \emph{\uniform}
iff for all  
$\dprb \in \dom(\dr)$ and 
$\act_1, \act_2, \actb_1, \actb_2$ acts of $\dprb$ such that
$\act_i \ind_\dprb \actb_i$, 
\commentout{
$$
\begin{array}{l}
\mbox{
$\act_1 \eleq_{\dr(\dprb)} \act_3$ iff $\act_2 \eleq_{\dr(\dprb)} \act_3$} \mbox{and}\\
\mbox{$\act_3 \eleq_{\dr(\dprb)} \act_1$ iff 
$\act_3 \eleq_{\dr(\dprb)} \act_2$}.
\end{array}$$
}%
\commentout{
\bit
\bl 
$\act_1 \eleq_{\dr(\dprb)} \act_3$ iff $\act_2 \eleq_{\dr(\dprb)} \act_3$ and \\
$\act_3 \eleq_{\dr(\dprb)} \act_1$ iff 
$\act_3 \eleq_{\dr(\dprb)} \act_2$.
\eit
}%
\[
\act_1 \leqRdp{\dprb} \act_2
\mbox{ iff }
\actb_1 \leqRdp{\dprb} \actb_2.
\]
Intuitively, we can think of \urvs\ and \ulotteries\ as descriptions of what an
act $\act$ does in terms of the tastes (and beliefs) of the \dmaker\@.  If $\dr$ is
\uniform, we can view $\dr$ as relating the acts indirectly by relating their
descriptions.

As the following theorem shows, all \uniform\ \drules\ 
that respects utility
have \gseu\
\representations.

\commentout{
Clearly \gseu\ is uniform, so a decision rule that is not uniform
cannot be represented by 
\gseu\@.
However, as the following result
shows, this is the only 
condition other than respecting utility that a  decision rule must
satisfy in order to be represented by 
\gseu\@.
}%

\newcounter{repSecNum}
\newcounter{repThmNum}
\setcounter{repSecNum}{\value{section}}
\setcounter{repThmNum}{\value{THEOREM}}
\thm\label{represent}  
\commentout{  
If
$\dr$ is a decision rule that
respects utility,\index{decision rule!respects utility} 
then $\dr$ is 
uniform iff $\dr$ can be represented by 
\gseu\@.
}%
For all 
\drules\ $\dr$, 
$\dr$ has a  \gseu\ \representation\ 
iff 
$\dr$ 
is \uniform\ and 
$\dr$ respects utility.
\ethm
\prf See the appendix.
\eprf

\commentout{
note that it is trivial to check that 
if $\dr$ has a \gseu\ \representation\ then $\dr$ respects utility. 
So suppose that $\dr$ 
respects utility but 
is not \uniform.
We need to show that $\dr$ has no \gseu\ \representation.
Since $\dr$ is not
\uniform, there exists some 
\commentout{
$\dprb_0 = ((\Act, \Stt, \Csq), \udom, \utf) \in \dom(\dr)$ 
and $\act_1, \act_2, \actb \in \Act$ such that
$\act_1 \ind_{\dprb_0} \act_2$ and either
}%
$\dprb_0 \in \dom(\dr)$ 
and acts
$\act_1,\act_2,\actb_1,\actb_2$ of $\dprb$ such that
$\act_i \ind_{\dprb_0} \actb_i$ and
\commentout{
\bit
\bl
$\act_1 \leqRdp{\dprb} \actb$ iff $\act_2 \not\leqRdp{\dprb} \actb$ or  \\
$\actb \leqRdp{\dprb} \act_1$ iff $\actb \not\leqRdp{\dprb} \act_2$.
\eit
}%
\begin{center}
$\act_1 \leqRdp{\dprb_0} \act_2$
\mbox{ iff }
$\actb_1 \not\leqRdp{\dprb_0} \actb_2$.
\end{center}
Since 
$\act_i \ind_{\dprb_0} \actb_i$, $\utf_{\act_i} = \utf_{\actb_i}$ (or
$\lto^{\plf,\utf}_{\act_i} =  \lto^{\plf,\utf}_{\actb_i}$), where
$\utf$ is the \utfn\ of $\dprb_0$ (and $\plf$ is the \plmsr\ of $\dprb$ in the
\plstic\ case). 
\commentout{
It is
clear that given any \plstic\ \dprob\ 
$\dprb \cong \dprb_0$, say 
$\dprb = (\ACT, \edom, \utf, \plf)$, where
$\edom = (\udom, \pdom, \eudom, \ep, \et)$, we must have
$\eplut{\act_1} = \eplut{\act_2}$.  
}%
It is easy to check that for all $(\ACT, \edom, \utf, \plf)\cong \dprb_0$,
$\eplut{\act_i} = \eplut{\actb_i}$.
Thus, for all $\dprb \cong \dprb_0$,
\commentout{
\bit
\bl
$\act_1 \leqRPdp{\mr{\gseu}}{\dprb} \actb$ iff $\act_2 \leqRPdp{\mr{\gseu}}{\dprb} \actb$ and \\
$\actb \leqRPdp{\mr{\gseu}}{\dprb} \act_1$ iff $\actb \leqRPdp{\mr{\gseu}}{\dprb} \act_2$.
\eit
}%
\begin{center}
$\act_1 \leqRPdp{\mr{\gseu}}{\dprb} \act_2$
\mbox{ iff }
$\actb_1 \leqRPdp{\mr{\gseu}}{\dprb} \actb_2$.
\end{center}
So $\mr{\gseu}(\dprb) \neq \dr(\dprb_0)$  for all $\dprb \cong \dprb_0$.
Thus
$\dr$ has no \gseu\ \representation.  
\eprf
}%

Most of the decision rules 
we
have discussed are 
uniform.  However, \ceu\ is not, 
as the following example shows:

\xam\label{xam:beldr}
Let
$\dprb_* = ((\Act, \Stt, \Csq), \sedom, \utf, \Bel),$ where 
\begin{itemize}
\item $\Act = \{\act_1, \act_2\}$; 
$\Stt = \{\stt_1, \stt_2, \stt_3\}$;
$\Csq = \{1,2,3\}$;
\item $\utf(j) = j,$ for $j = 1, 2, 3$;
\item $\act_1(\stt_j) = j$ and $\act_2(\stt_j) = 4-j,$ for $j = 1, 2, 3$; and
\item $\Bel$ is the belief function such that
$\Bel(X) = 1$ if $\{\stt_1,\stt_2\} \sbset X$ and
$\Bel(X) = 0$ otherwise.
\end{itemize}
\commentout{
Since $\utf_{\act_i}^{-1}(j)$ is a singleton, $\act_1 \ind_{\dprb_*} \act_2$, 
so
}%
Since $\utf_{\act_i}^{-1}(j)$ is a singleton, 
$\Bel(\utf_{\act_i}^{-1}(j)) = 0$ for $i = 1, 2$ and $j = 1, 2, 3$;
thus $\act_1 \ind_{\dprb_*} \act_2$.
On the other hand, by definition,
\commentout{
$$
\E_{\Bel}(\utf_{\act_1}) = 1 + (2-1)\Bel(\{w_2,w_3\}) + (3-2) \Bel(\{w_3\})
= 1,$$
while 
$$\E_{\Bel}(\utf_{\act_2}) = 1 + (2-1)\Bel(\{w_1,w_2\}) + (3-2) \Bel(\{w_1\})
= 2.$$
}%
\[
\E_{\Bel}(\utf_{\act_1}) = 1 + \Bel(X_{23})(2-1) + \Bel(X_3)(3-2)
= 1,
\]
while
\[
\E_{\Bel}(\utf_{\act_2}) = 1 + \Bel(X_{12})(2-1) + \Bel(X_1)(3-2)
= 2,
\]
where $X_{ij} = \{\stt_i,\stt_j\}$ and $X_k = \{\stt_k\}$.
It follows that \ceu\ is not uniform, and so
\commentout{
cannot be represented by 
\gseu\@.
}%
has no \gseu\ \representation.
\exam

The 
reader may have noticed an incongruity here.
Example~\ref{xam:mmeu} shows that 
\mmeu\ has a \gseu\ \representation;
moreover, as shown earlier, \mmeu\ produces essentially the same order
on 
acts
as 
\ceu\
restricted to belief functions.
\commentout{
However, \gseu\ cannot represent 
\ceu\@.
}%
However, \ceu\ has no \gseu\ \representation.
There is no contradiction to
Theorem~\ref{represent} 
here:
\commentout{
If $\dprb$ is the decision problem in
Example~\ref{xam:beldr}, then
there is no decision problem 
$\dprb'$ that is congruent to $\dprb$ such that
$\mr{\gseu}(\dprb') = \mr{\ceu}(\dprb)$.
}%
There is no \dprob\ $\dprb$ such that 
$\dprb \cong \dprb_*$ (from \mxam{xam:beldr}) and
$\mr{\gseu}(\dprb) = \mr{\ceu}(\dprb_*)$.
However,
$\mr{\gseu}((\Act,\Stt,\Csq),\edom_{\P_\Bel},\utf,\Pl_{\P_{\Bel}}) = \mr{\ceu}(\dprb_*)$. 
Of course, 
$((\Act,\Stt,\Csq),\edom_{\P_\Bel},\utf,\Pl_{\P_{\Bel}}) \not\cong \dprb_*$;
$\Pl_{\P_{\Bel}}$ and $\Bel$ 
are \emph{not} the same, and they in fact represent related but different beliefs.
(It is easy to show that sets are 
partially preordered by $\plf_{\P_{\Bel}}$ but
totally preordered by $\Bel$.)

The key reason that \gseu\ cannot \represent\ \nonuniform\ \drules\ is
because they do not respect the indistinguishability relations imposed by the
\utfn\ (and the \plmsr).
Recall that we require that $\trans(\dprb) \cong \dprb$ because
we want a user of one \drules\ to appear as if she were using another, without
pretending that she has different tastes (and beliefs).  So we want $\trans$ to
preserve the tastes (and beliefs) of its input.
\commentout{
For a decision rule $\dr_1$ to represent a rule $\dr_2$, there
must be a decision problem transformation $\tau$ such that
$\dr_1(\tau(\dprb)) = \dr_2(\dprb)$ for every decision problem $\dprb$
in the domain of $\dr_2$.  Since $\tau(\dprb)$ and $\dprb$ are
congruent, they agree on the tastes and (if they are both
plausibilistic) the beliefs of the 
DM\@.
We now weaken this
requirement somewhat, and consider transformations that preserve
an important aspect of a DM's tastes (and beliefs), while not
requiring them to stay the same.
}%

\commentout{
The key reason that \gseu\ cannot \represent\ \nonuniform\ \drules\ is
because they do not respect the indistinguishability relation imposed by the
\utfn\ (and the \plmsr).
Recall that we require that $\trans(\dprb) \cong \dprb$ because
we want a user of one \drules\ to appear as if she were using another, without
pretending that she has different tastes (and beliefs).  So we want $\trans$ to
preserve the tastes (and beliefs) of its input.
Note that
$\trans(\dprb) \cong \dprb$ means that the \utfn\ (and \plmsr) in
$\trans(\dprb)$ are the same as those in $\dprb$; this is clearly sufficient to
preserve the tastes (and beliefs), but it is arguably not necessary, since different
\utfns\ (and \plmsrs) could represent the same tastes (and beliefs).
We define below what it means for different \utfns\ (and \plmsrs) to
\represent\ the same tastes (and beliefs). It would suffice then to require
that the \utfn\ (and \plmsr) in $\trans(\dprb)$ represent(s) the same tastes (and
beliefs) as the one(s) in $\dprb$. 
}%

\commentout{
Note that the key reason that \gseu\ cannot \represent\ \nonuniform\ \drules\ is
because they do not respect the indistinguishability relation imposed by the
\utfn\ (and the \plmsr).  As we show below, if we are allowed to modify 
the \representations\ of the tastes (and beliefs), then we can essentially
express all \drules\ in terms of \gseu\@. 

Also, the key reason that \gseu\ cannot \represent\ \nonuniform\ \drules\ is
because they do not respect the indistinguishability relation imposed by the
\utfn\ (and the \plmsr).  

It turns out that if we allow $\trans$ to modify the \utfn\ (and
\plmsr), then we can express almost all \drules\ in terms of \gseu; the trick
is to modify the \utfn\ (and \plmsr) in a way that somehow leaves the tastes (and
beliefs) the same.
To do this, we need to first define what it means for two
\utfns\ to \represent\ the same tastes.

So if we somehow modify the \utfn\ (and the \plmsr), while leaving
the underlying tastes (and beliefs) the ``same'' (in the sense defined below),
We now show that if we allow $\trans$ to modify the \emph{representation} of
the tastes (and beliefs), while leaving the tastes (and beliefs) the ``same''
(in the sense defined below), then almost all \drules\ can be expressed in
terms of \gseu\@.  To do this, we first define what it means for two \utfns\
to \represent\ the same tastes.
}%

\commentout{
One obvious
necessary condition for one \utfn\ $\utf_1 : \Csq \rar \udom_1$ to \represent\
the same tastes as $\utf_2 : \Csq \rar \udom_2$ is
$\utf_1(\csq_1) \eleq_{\udom_1} \utf_1(\csq_2)$  iff 
$\utf_2(\csq_1) \eleq_{\udom_2} \utf_2(\csq_2)$
for all $\csq_1, \csq_2 \in \Csq$. Whether this is enough depends on what
other functions or relations are present.  While we do not assume that a
\utstr\ $\udom$
has any functions or relations besides $\eleq_\udom$, 
we do not insist that no such functions or
relations exist either.  Indeed, the standard \utstr, $\realNum$, comes with
addition ($+$) and multiplication ($\times$), for example.  
So, in the case of
$\realNum$, it 
might be that 
$\utf_2(\csq_1) \leq (\utf_2(\csq_2) + \utf_2(\csq_3)) \times \utf_2(\csq_4)$;
should we require that this holds of $\utf_1$ as well?
Certainly it would not hurt to make such a requirement, though it is arguable
that some properties of $\utf_2$ are not salient features of the
underlying tastes, so there is no need to impose such requirements.
However, we do not want to arbitrarily declare what is or is not salient
(especially since we consider general \utstrs), so we
want to make sure that $\utf_1$ satisfies as many of the properties that 
$\utf_2$ satisfies as possible. 
should we
insist that 
\[
\utf_1(\csq_1) \leq (\utf_1(\csq_2) + \utf_1(\csq_3)) \times \utf_1(\csq_4)
\] as well?
Certainly it would not hurt to make such a requirement, though it is arguable
that some properties of $\utf_2$ are not salient features of the
underlying tastes, so there is no need to impose such requirements.
However, we do not want to arbitrarily declare what is or is not salient
(especially since we consider general \utstrs), so we
want to make sure that $\utf_1$ satisfies as many of the properties that 
$\utf_2$ satisfies as possible. 
}%

\commentout{
One way to ensure that $\utf_1$ is essentially the same as $\utf_2$ is to 
require that $\udom_1$ and
$\udom_2$ are isomorphic, and that there is an isomorphism 
$I : \udom_1 \rar \udom_2$ such that $I(\utf_1(\csq)) = \utf_2(\csq)$ for all 
$\csq \in \Csq$.  However, it is not hard to see that this is too strong, since
pairs indistinguishable under $\utf_2$ remain indistinguishable under
$\utf_1$. To allow $\utf_1$ to distinguish more pairs than $\utf_2$, 
we remove the requirement that the correspondence between
$\udom_1$ and $\udom_2$ is \mbox{1-1}.  We say that 
$\utf_1 : \Csq \rar \udom_1$ \emph{extends}
$\utf_2 : \Csq \rar \udom_2$
iff
there is an onto homomorphism $h : \udom_1 \rar \udom_2$ such that
$h(\utf_1(\csq)) = \utf_2(\csq)$ for all $\csq \in \Csq$.
Similarly, we say that $\plf_1 : 2^\Stt \rar \pdom_1$ 
\emph{extends}
$\plf_2 : 2^\Stt \rar \pdom_2$
iff
there is an onto homomorphism $h : \pdom_1 \rar \pdom_2$ such that
$h(\plf_1(X)) = \plf_2(X)$ for all $X \sbset \Stt$.
(We assume the reader is familiar with homomorphisms.)
We say that $\dprb_1$ \emph{extends} $\dprb_2$, denoted
$\dprb_1 \ext \dprb_2$, iff the \utfn\ (and \plmsr) of $\dprb_1$ extend(s) 
the \utfn\ (and \plmsr) of $\dprb_2$.
}%

\commentout{
We say that 
$\trans$ is a \emph{homomorphic} $\dr_1$ \representation\ of $\dr_2$ iff
$\dom(\trans) = \dom(\dr_2)$ and
for all $\dprb \in \dom(\dr_2)$,
\[
\trans(\dprb) \ext \dprb \mbox{ and }
\dr_1(\trans(\dprb)) = \dr_2(\dprb).
\]
}%

There is a long-standing debate in the 
decision-theory literature 
as to
whether preferences  should be 
regarded as
\emph{ordinal} or 
\emph{cardinal}.  If they are ordinal, then all that matters is their order.
If they are cardinal, then it should be meaningful to talk about the
\emph{differences} between 
preferences---that is, how much more a
DM prefers one consequence to another.  
Similarly, if
representations
of likelihood are taken to be ordinal, then all that matters
is whether one event is more likely than another.  
As we show below, if we require only that $\trans(\dprb)$ and $\dprb$ describe the
same ordinal tastes (and beliefs), then we can in fact express almost all
\drules, including \ceu, in terms of \gseu\@.

Two \utfns\
$\utf_1 : \Csq \rar \udom_1$ and $\utf_2 : \Csq \rar \udom_2$ 
\emph{represent the same ordinal tastes}
if
for all $c_1, c_2 \in \Csq$,  
\[
\utf_1(c_1) \eleq_{\udom_1}  \utf_1(c_2) \mbox{ iff }
\utf_2(c_1) \eleq_{\udom_2} \utf_2(c_2).  
\]
Similarly, two
plausibility measures 
$\plf_1 : 2^\Stt \rar \pdom_1$ and
$\plf_2 : 2^\Stt \rar \pdom_2$ 
\emph{represent the same ordinal beliefs}
iff for all $X, Y \sbset \Stt$, 
\[
\plf_1(X) \pleq_{\pdom_1} \plf_1(Y) \mbox{ iff }
\plf_2(X) \pleq_{\pdom_2} \plf_2(Y).
\]
Finally, two decision problems $\dprb_1$ and $\dprb_2$ are 
\emph{similar}, denoted $\dprb_1 \ssim \dprb_2$, 
iff they involve the same
decision situations, their utility functions represent the same ordinal
tastes, and their plausibility measures represent the same ordinal
beliefs.  Note that 
$\dprb_1 \cong \dprb_2$ implies $\dprb_1 \ssim \dprb_2$, but the converse is
false in general.
A \druletr\ $\trans$ is an \emph{ordinal $\dr_1$ \representation\ of $\dr_2$} iff
$\dom(\trans) = \dom(\dr_2)$ and 
for all $\dprb \in \dom(\dr_2)$,
\bit
\bl $\trans(\dprb) \ssim \dprb$ and 
\bl $\dr_1(\trans(\dprb)) = \dr_2(\dprb)$.
\eit
\commentout{
Decision rule $\dr_1$ can
\emph{emulate} decision rule $\dr_2$
iff there exists a mapping 
$\tau$\index{decision problem transformation}\glossary{\glostau} from
decision problems to similar decision problems such 
that, for all decision problems $\dprb$ in the domain of $\dr_2$, 
the decision problem $\tau(\dprb)$ is in the domain of $\dr_1$ and 
$\dr_1(\tau(\dprb)) = \dr_2(\dprb)$.  
Thus, the definition of emulation is just like that of representation,
except that $\tau(\dprb)$ is now required only to be similar to $\dprb$,
not congruent.
}%

We want to show next that almost all \drules\ have an ordinal \gseu\
\representation. 
Doing so involves one more subtlety. 
Up to now, we have assumed that 
\plstrs\ are partially ordered.
This implies that two \plmsrs\ that \represent\ the same ordinal beliefs
necessarily induce the same indistinguishability relation (because of
anti\-symmetry). 
Thus, in order to distinguish sets that have equivalent plausibilities
when computing expected utility using $\et$ and $\et$, we
need to allow \plstrs\ to be partially
\emph{preordered}.
So, for this result, we assume that
$\eleq_\pdom$ is a reflexive and
transitive relation that is not necessarily anti\-symmetric (\ie\ we could have
that $p_1 \eleq_\pdom p_2$ and $p_2 \eleq_\pdom p_1$ but $p_1 \neq p_2$).
\newcounter{repoSecNum}
\newcounter{repoThmNum}
\setcounter{repoSecNum}{\value{section}}
\setcounter{repoThmNum}{\value{THEOREM}}
\thm\label{represent1}  
\commentout{
If  $\dr$ is a decision
rule that respects utility\index{decision rule!respects utility}, then \gseu\ can emulate 
$\dr$.
}%
A \drule\ $\dr$ has an ordinal \gseu\ \representation\ 
iff $\dr$ weakly respects utility.
\ethm
\prf
See the appendix. \eprf

Theorem~\ref{represent1} shows that \gseu\ can 
ordinally represent essentially all
decision rules.  Thus, there is a sense in which \gseu\ can be viewed
as a universal decision rule.

\commentout{
(We remark that although we have focused here on alternatives that are
acts, in the sense of Savage, that is, functions from states to
consequences, it is not hard to show---and we do in the full paper---that
the same results hold if alternatives are taken to be horse lotteries,
in the sense of Anscombe and Aumann 
\citeyear{AA1963}.)
}%

\section{Related Frameworks}
\label{sec:lottery}

\commentout{
Readers familiar with decision theory will note that so far we have formalized
the notions of \dsitu, \dprob, and \drule\ 
}%
Note that so far we have worked
exclusively in the \emph{act framework} used by Savage~\citeyear{Savage}. 
There are some other well-known frameworks in the decision-theory literature;
perhaps the two best-known such frameworks are the
\emph{lottery framework} introduced by \vonNM~\citeyear{vNM1947}, and
Anscombe and Aumann's \citeyear{AA1963} {\em  horse lotteries}, which
can be viewed as a combination of the act and lottery frameworks.
Since our goal is to provide a single framework for almost all of decision
theory, 
in this section we briefly discuss how the act framework can model these,
in much the same way as
Turing machines can model other notions of computation.
We begin with the lottery framework.

\subsection{The Lottery Framework}

As the name suggests, the alternatives in the  lottery framework are lotteries, or
probability distributions over \consequences.
Standard lotteries are functions of the form $\lto : \Csq \rar [0,1]$
such that $\sum_{\csq \in \Csq} \lto(\csq) = 1$.
A standard lottery is \emph{simple} iff 
$\{ \csq \st \lto(\csq) > 0\}$, which is 
typically called the \emph{support} of $\lto$ and is denoted $\ssupp(\lto)$, 
is finite.  
Note that the support of a  standard lottery is nonempty.

In general, we want to allow 
assignments of plausibilities to 
sets of 
\consequences.
Given a set of \consequences\ $\Csq$ and a \plstr\ $\pdom$, a
\emph{lottery} is a 
plausibility measure $\lto: 2^Q \rar \pldom$, where $Q$ is a 
nonempty subset of $\Csq$.
We denote $Q$ as $\supp(\lto)$.
In the standard case, we take $Q$ to consist of those consequences $\csq$
such that $\lto(\csq) > 0$, so
$\sum_{\csq \in \supp(\lto)} \lto(\csq) = 1$.
\commentout{
The reason that we do not
take lotteries to be functions of the form
$\lambda : \Csq \rar \pdom$ and just identify a function
$\lto : X \rar \pdom$ with some $\lambda$
let 
$X = \{ \csq \st \lambda(\csq) \egr \bottom \}$ 
}%
We say that $\lto$ is \emph{degenerate} if $\abs{\supp(\lto)} = 1$,
and we say that a lottery $\lto$ is \emph{simple} iff $\supp(\lto)$ is finite. 
Just as we focus on 
simple acts, we 
focus on simple lotteries 
(as did \vonNM~\citeyear{\vNM1947}).
Though lotteries are functions that assign plausibility values to consequences,
we follow a common convention in the literature that lists plausibilities
first (\eg\ see \cite{SW1997,WZ2002}). 
So
$\{(p_1,\csq_1), \ldots, (p_n,\csq_n)\}$ denotes the lottery $\lto$ such that
$\supp(\lto) = \{\csq_1, \ldots, \csq_n\}$ and 
$\lto(\csq_i) = p_i$.  (Note that this is the reverse of the usual notation for
functions.)

Many notions we defined in the act framework have counterparts in the lottery
framework.  For example, the counterpart of a \dsitu\ is a lottery \dsitu.
Formally, a 
\emph{lottery \dsitu} is a tuple $\LTO = (\Lto,\Csq,\pdom)$, where
\bit
\bl $\Csq$ is a set of \consequences, 
\bl $\pdom$ is a \plstr, and
\bl $\Lto$ is a 
(nonempty)
set of simple lotteries 
over $\Csq$.
\eit
Note that a lottery 
\dsitu\ does not contain any information about the tastes of the \dmaker\@.  A
lottery \dprob\ is essentially a lottery \dsitu\ together with information
about the tastes of the \dmaker\@.  Formally, a \emph{lottery \dprob} is a
tuple $(\LTO,\edom,\utf)$, where
\bit
\bl $\LTO = (\Lto,\Csq,\pdom)$ is a lottery \dsitu,
\bl $\edom = (\udom,\pdom,\eudom,\ep,\et)$ is an \expstr, and
\bl $\utf : \Csq \rar \udom$ is a \utfn.
\eit
Note that the \plstr\ of the \expstr\ is the same as the \plstr\ of the lottery
\dsitu.  

A \emph{standard lottery \dprob} is a lottery \dprob\ with the standard
\expstr; 
these are the ones that are studied extensively in the literature.
Perhaps the best-known lottery
\drule\ is \vonNM's expected utility rule: choosing the lottery
that maximizes expected 
utility---that is, 
choosing the lottery $\lto$
that 
maximizes
\beqn
\label{eqn:lotteryEU-std}
\E_{\lto}(\utf) = \sum_{\csq \in \supp(\lto)} \lto(\csq) \times \utf(\csq).
\eeqn
As in the act framework, we can generalize 
(\ref{eqn:lotteryEU-std}) to arbitrary \expstrs:
\beqn
\label{eqn:lotteryEU}
\E_{\lto,\edom}(\utf) = \EP_{\csq \in \supp(\lto)} \lto(\csq) \et \utf(\csq).
\eeqn
Some 
other 
well known lottery \drules\ include
disappointment aversion (DA)~\cite{Gul1991},
rank-dependent expected utility (RDEU)~\cite{Quiggin93,Yaari1987},
and cumulative prospect theory (CPT)~\cite{WZ2002}. 

Our goal in this section is to show that the act framework can model the
lottery framework, so that it suffices for the rest of the \paper\ to focus
only on the act framework.  To 
facilitate 
this, we 
introduce
one other notion in the act
framework.
A \emph{\plstic\ \dsitu} is a tuple $(\ACT,\pdom,\plf)$, where
\bit
\bl $\ACT = (\Act,\Stt,\Csq)$ is a \dsitu,
\bl $\pdom$ is a \plstr, and
\bl $\plf : 2^\Stt \rar \pdom$ is a \plmsr.
\eit
Like a lottery \dsitu, a \plstic\ \dsitu\ describes the beliefs but not the
tastes of the \dmaker\@.  The difference is, of course, that the belief of the
\dmaker\ is described by a single \plmsr\ as opposed to a set of lotteries.
Note that a \plstic\ \dprob\ is essentially a \plstic\ \dsitu\ together with a
\utfn. 

Given a \plstic\ \dsitu\
$\STT = ((\Act,\Stt,\Csq),\pdom,\plf)$, each $\act \in \Act$ 
induces a lottery
$\lto^\plf_\act$ 
as follows:  
$\supp(\lto^\plf_\act) = \ran(\act)$ and 
$\lto^\plf_\act(Y) = \plf(\act^{-1}(Y))$ for 
$Y \sbset \ran(\act)$. 
Note that if
$\act$ is simple, then $\lto^\plf_\act$ is also simple.
We say that a \plstic\ \dsitu\ $\STT$ \emph{induces} 
the lottery \dsitu\ 
$\LTO_\STT = (\{\lto^\plf_\act \st \act \in \Act\},\Csq,\pdom)$.
This mapping from plausibilistic \dsitus\ to lottery \dsitus\ is clearly
not \mbox{1-1}.  It is possible to have $\STT_0 \neq \STT_1$ but 
$\LTO_{\STT_0} = \LTO_{\STT_1}$, since different acts could induce the same
lotteries (in fact, $\STT_0$ and $\STT_1$ may even involve different sets of
states). 
However, as the following result shows, the mapping from plausibilistic
\dsitus\ to lottery \dsitus\ is onto.
\newcounter{redSecNum}
\newcounter{redThmNum}
\setcounter{redSecNum}{\value{section}}
\setcounter{redThmNum}{\value{THEOREM}}
\pro
\label{pro:stt-induce-lto2}
Every lottery \dsitu\ $\LTO = (\Lto,\Csq,\pdom)$ is induced by some 
\plstic\ \dsitu\ 
$\STT_\LTO$.
\epro
\prf See the appendix.
\eprf

\commentout{
Recall that the reason that we did not let
$\dom(\lto) = \supp(\lto)$ is because 
plausibility distributions in general, unlike probability distributions, are
not determined by its behavior on its support.
In order to define $\lto^\utf$, for example, we needed to know in general the 
plausibility of sets of consequences, since $\utf$ might not be 1-1.
However, we are often interested in the function from $\Csq$ (or $\udom$, in
the case of utility lotteries) to $\pdom$ induced
by $\lto$, denoted $\atm{\lto}$, defined as 
\[
\atm{\lto}(\csq) = \left\{
\begin{array}{ll}
\bottom & \mbox{for } \csq \notin \supp(\lto) \mbox{ and}\\
\lto(\{\csq\}) & \mbox{for } \csq \in \supp(\lto). 
\end{array}
\right.
\]
To avoid cluttering the presentation, for the rest of
this \paper, unless otherwise stated, we will identify $\lto$ with
$\atm{\lto}$ and $\lto^{\utf}$ with $\atm{\lto^{\utf}}$.
}%

\cor Every \prefrel\ in the lottery framework can be modeled by a
\prefrel\ in the act framework.
\ecor
\prf 
Let $\STT = ((\Act,\Stt,\Csq),\pdom,\plf)$ be a \plstic\ \dsitu\ and let 
$\LTO = (\Lto,\Csq,\pdom)$ be the lottery \dsitu\ it induces.  Note that every
\prefrel\ $\eleq_\Lto$ on the lotteries in $\Lto$ induces a \prefrel\ $\leqact$
on the acts in $\Act$ as follows:
\[
\act_1 \leqact \act_2 \mbox{ iff } 
\lto^\plf_{\act_1} \eleq_\Lto \lto^\plf_{\act_2}.
\]
In other words, $\leqact$ relates acts
by the way $\eleq_\Lto$ relates the lotteries they induce.
Since every lottery \dsitu\ is induced by some \plstic\ \dsitu\
(by \mpro{pro:stt-induce-lto2}), every \prefrel\ in the lottery framework can be
modeled in the act framework.
\eprf

Note that an arbitrary \prefrel\ $\leqact$ on the acts in $\Act$ does not
correspond to a \prefrel\ $\eleq_\Lto$ on the lotteries in $\Lto$ in general,
since $\leqact$ could treat acts that induce the same lottery differently.
In order for $\leqact$ to correspond to some $\eleq_\Lto$, it must be
\emph{lottery-\uniform}, in the sense that, 
for all
acts $\act_1, \act_2, \actb$, if 
$\lto^\plf_{\act_1} = \lto^\plf_{\act_2}$, then  
\[
\act_1 \leqact \actb \mbox{ iff } 
\act_2 \leqact \actb \mbox{ and }
\actb \leqact \act_1 \mbox{ iff } 
\actb \leqact \act_2.
\]
It is not hard to see 
that lottery-\uniform\ \prefrels\ on acts are exactly those induced by
\prefrels\ on the lotteries.  
It is also not hard to see that not all \prefrels\ on acts are
lottery-\uniform.  So, some preferences that can be described by relating acts
in a \plstic\ \dsitu\ $\STT$ cannot be described by relating the lotteries in the
lottery \dsitu\ $\STT$ induces.

Turning now to \dprobs\ and \drules,
we say that a \plstic\ \dprob\ $\dprb_{\Act} = (\ACT,\edom,\utf,\plf)$
\emph{induces} the lottery \dprob\ $\dprb_{\Lto} = (\LTO,\edom,\utf)$,
where $\LTO$ is the lottery \dsitu\ induced by the 
the \plstic\ \dsitu\ of $\dprb_{\Act}$, $(\ACT,\pdom,\plf)$. 
Since every \plstic\ \dprob\ induces a unique lottery \dprob, every 
lottery
\drule\ $\dr_{\Lto}$ induces
a 
\plstic\ \drule\ $\dr_{\Act}$
as
follows:
\[
\act_1 \eleq_{\dr_{\Act}(\dprb_{\Act})} \act_2 \mbox{ iff }
\lto^\plf_{\act_1} \eleq_{\dr_{\Lto}(\dprb_{\Lto})} \lto^\plf_{\act_2},
\]
where $\dprb_{\Lto}$ is the lottery \dprob\ induced by $\dprb_{\Act}$.
Basically,  $\dr_{\Act}$ relates acts 
by relating the lotteries they induce using $\dr_\Lto$.
The domain of $\dr_\Act$ is 
$\{ \dprb_\Act \st \dprb_\Act$ induces 
some $\dprb_\Lto \in \dom(\dr_\Lto) \}$.  
Thus every 
lottery \drule\ can be modeled by a 
\plstic\ \drule.
\commentout{
It follows that our results about \gseu\ representing (almost) all
\drules\ apply immediately to lottery \drules; in addition, it follows
that the results of our companion paper showing that every preference
order on acts can be represented by \gseu\ also show that every
preference order on lotteries can be represented by \gseu.
}%
Using these observations, it is not hard to show that
most of our results in previous sections also hold in the lottery
framework.  For example, it is easy to show that \gseu\ when applied to
lotteries yields a lottery \drule\ that can \represent\ 
all \prefrel\ on lotteries and 
almost all lottery
\drules.  (More precisely, lottery \gseu\ can \represent\ all \uniform\ lottery
\drules, where the notion of uniformity is completely analogous to the one
presented in \msec{sec:drule}.)
However, as we have just seen, all lottery \drules\ can be modeled by
\plstic\ \drules.  Thus it suffices, from a technical perspective, to
focus exclusively on the act framework, as we have done in this \paper, when
considering the foundations of decision theory.

\commentout{
What we have shown is that given any set of
objective plausibility distributions, we can model it using 
a subjective \plmsr\ (

Before we end this section, we would like to make a brief comment about a hybrid
framework
due to Anscombe and Aumann~\cite{AA1963},
which we refer to as the \emph{AA framework}.  
}%

\subsection{The \AAu\ Framework}

Anscombe and Aumann \citeyear{AA1963} define a framework that is 
essentially a combination of the act framework and the lottery
framework: basically, it takes the consequences in the act framework and
replaces them by lotteries, so acts (also known as  
\emph{horse lotteries}) map  
states to lotteries (also known as \emph{roulette lotteries})
The probabilities 
that
the roulette lotteries assign to consequences are typically
regarded as ``objective'' (in the sense that they are 
determined by the properties of the devices, such as fair
coins or unloaded dice, used to generate them),
while the probabilities (if any) associated with the sets of
states are regarded, as in the act framework, as ``subjective'' (in the sense
that these
describe
the beliefs of the \dmaker)\@. 

We can formalize the AA framework in much the same way we formalized 
the act and lottery frameworks.
\commentout{
We first present a formalization that is quite general, in the sense that we do
not restrict what kind of domains could be involved;
}%
As usual, we begin with \dsitus.
An \emph{AA \dsitu} is a tuple
$\HLT = (\Hlt,\Stt,\LTO)$, where
\bit
\bl $\Stt$ is a set of \ssofn,
\bl $\LTO = (\Lto, \Csq, \pdom)$ is a lottery \dsitu, 
and
\bl $\Hlt$ is a nonempty set of horse lotteries (\ie\ a nonempty subset of
$\Lto^\Stt$).
\eit
A \emph{\nonplstic\ AA \dprob} is a tuple
$(\HLT,\widehat{\edom},\utf)$, where
\bit
\bl $\HLT = (\Hlt,\Stt,(\Lto,\Csq,\widehat{\pdom}))$ is an AA \dsitu,
\bl $\widehat{\edom} = (\widehat{\udom},\widehat{\pdom},\widehat{\eudom},\widehat{\ep},\widehat{\et})$
is an \expstr, 
and 
\bl 
$\utf : \Csq \rar \widehat{\udom}$ is a \utfn.
\eit
Finally, a \emph{\plstic\ AA \dprob} is a tuple
$(\HLT,\widehat{\edom},\utf,\edom,\plf)$, where
\bit
\bl $\HLT = (\Hlt,\Stt,(\Lto,\Csq,\widehat{\pdom}))$ is an AA \dsitu, 
\bl $\widehat{\edom} =
(\widehat{\udom},\widehat{\pdom},\widehat{\eudom},\widehat{\ep}, \widehat{\et})$ is an \expstr\
(for roulette lotteries), 
\bl $\utf : \Csq \rar \widehat{\udom}$ is a \utfn, 
\bl $\edom = (\widehat{\eudom},\pdom,\eudom,\ep,\et)$ is an \expstr\ 
(for horse lotteries),
and
\bl $\plf : 2^\Stt \rar \pdom$ is a \plmsr.
\eit
We need two \expstrs, since in general 
the objective uncertainties  and subjective uncertainties could be expressed in
different languages.  Note that 
the \utstr\ of $\edom$ is the \valstr\ of $\widehat{\edom}$, so
the expected utility values with respect to the roulette lotteries are
the utility values 
for $\edom$.  
\commentout{
We can think of a horse lottery as an act 
whose consequences are 
roulette lotteries.  
}%
While the formalization above is somewhat involved, in the standard setting,
$\widehat{\edom} = \sedom$, and for the \plstic\ case, $\edom = \sedom$ as
well.

In the standard setting, it is  quite common to have the \utfn\ map
roulette lotteries, rather than just the (deterministic) consequences, to real
numbers---that is, the domain of $\utf$ is $\Lto$ rather than $\Csq$;
see, for example, \cite{AA1963,GS1989,Schmeidler89}. 
This is because a utility function $\utf$ defined on $\Csq$ can easily be
extended to $\Lto$ by taking $\utf(\lto) = E_\lto(\utf)$.
We can similarly extend $\utf$ to $\Lto$ in our framework, by taking
$\utf(\lto) = \E_{\lto,\widehat{\edom}}(\utf)$.  
Note that if $\lto$ is degenerate with $\supp(\lto) = \{\csq\}$, then
\[
\utf(\lto) = \E_{\lto,\widehat{\edom}}(\utf) = \top_{\widehat{\pdom}}
\mathop{\widehat{\et}} \utf(\csq) = \utf(\csq),
\]
as one would expect.

Once we extend $\utf$ to $\Lto$ and treat lotteries as consequences, we can
essentially view the AA framework as a special case of the act framework.
As usual,
a horse lottery $\hlt$ induces the random variable 
$\utf_\hlt : \Stt \rar \widehat{\eudom}$ as follows:
$\utf_\hlt(s) = \utf(\hlt(s))$.  
The expected utility of a horse lottery $\hlt$ is then
\[
\E_{\plf,\edom}(\utf_\hlt)
\bydef{=}
\EP_{x \in \ran(\utf_\hlt)}
\plf(\utf_\hlt^{-1}(x)) \et x.
\]
Thus, again, for the purpose of studying the foundations of
decision theory, it suffices to focus on the act framework, since all \drules\
in the AA framework can also be modeled by \drules\ in the act framework.

\commentout{
While the formalization above is somewhat involved, in the standard setting,
$\widehat{\edom} = \sedom$, and for the \plstic\ case, $\edom = \sedom$ as
well.  
}%
\commentout{
Results in the AA framework typically characterizes when it is possible
to represent a \prefrel\ $\eleq_\Hlt$ on horse lotteries using some \drule; for
example, 
\AAumann~\citeyear{AA1963} consider \seu, 
Gilboa and Schmeidler~\citeyear{GS1989} consider \mmeu, and
Schmeidler~\citeyear{Schmeidler89} considers \ceu\@.
}%
\commentout{

Unlike the case with the lottery framework, a \dprob\ in the act framework does
not induce a \dprob\ in the AA framework.  However, we can easily model
\dprobs\ in the AA framework in the act framework.
For ease of exposition, we identify $\csq \in \Csq$ with the 
degenerate
lottery $\{ (\top,\csq) \}$.

Note that the AA framework can be considered a generalization of the act
framework, since we can simply identify a deterministic consequence $\csq$ with
the degenerate lottery $\{(\top,\csq)\}$, and view any \dsitu\ as an AA \dsitu\ in
which the set $\Lto$ of roulette lotteries consists of only 
degenerate lotteries
(In the standard setting, however, the set of roulette lotteries is typically
assumed to be convex.)
}%

\commentout{
Though the AA framework seems more general than the act framework, it can
easily be modeled in the act framework.
For ease of exposition, we follow a common practice
of identifying $\csq \in \Csq$ with the degenerate lottery $\{ (\top,\csq) \}$
which has only $\csq$ in its domain \cite{GS1989,Schmeidler89}.  
Given a \nonplstic\ AA 
\dprob\ $\dprb_\Hlt = ((\Hlt,\Stt,(\Lto,\Csq,\widehat{\pdom})),\widehat{\edom},\utf)$,
where $\widehat{\edom} = (\widehat{\udom},\widehat{\pdom},\widehat{\eudom},\widehat{\ep},\widehat{\et})$, 
we can model it using 
$\dprb = ((\Hlt,\Stt, \Lto), \widehat{\eudom}, \E_{\utf,\widehat{\edom}})$. 
Note that $\Lto$ is the set of consequences,
$\eudom$  is the \utstr, and 
$\E_{\utf,\widehat{\edom}}$ 
is the \utfn.
Note also that 
$\E_{\utf,\widehat{\edom}} : \Lto \rar \widehat{\eudom}$ and 
$\E_{\utf,\widehat{\edom}}(\csq) = \top \widehat{\et} \utf(\csq) = \utf(\csq)$
for all $\csq \in \Csq$,
so 
$\E_{\utf,\widehat{\edom}}$ 
is basically an extension of
$\utf$ to the roulette lotteries.

The \plstic\ case can be handled in a similar way.  
Given a \plstic\ AA 
\dprob\ $\dprb_\Hlt = ((\Hlt,\Stt,(\Lto,\Csq,\pdom)),\widehat{\edom},\utf,\edom,\plf)$,
where $\widehat{\edom} = (\widehat{\udom},\widehat{\pdom},\widehat{\eudom},\widehat{\ep},\widehat{\et})$
and $\edom = (\widehat{\eudom},\pdom,\eudom,\ep,\et)$, 
we can model it using 
$\dprb = ((\Hlt,\Stt, \Lto), \edom, \E_{\utf,\widehat{\edom}}, \plf)$.
\commentout{
While we have not considered consequences with internal
structure before this, nothing in the formal framework excludes such possibilities.
Furthermore, it is quite common in the literature to take the domains of
\utfns\ to be the roulette lotteries when working in the AA
framework;\footnote{In fact, as Savage \citeyear{Savage} points out, it is actually
quite common in the economics literature (especially after \vonNM's results
became known)
to let the domain of \utfns\ be
(roulette) lotteries, which are considered \emph{risky} alternatives, rather
than just (deterministic) consequences, which are considered \emph{riskless} or
\emph{certain} alternatives.}
see, for example, \cite{AA1963,GS1989,Schmeidler89}. 
Also, even if we insist that consequences cannot be roulette lotteries, it is easy
to encode the roulette lotteries  somehow, and let the codes of roulette
lotteries be the consequences, and we let the utility of the codes be the
expected utility of the corresponding roulette lotteries;  all that would
happen is the presentation gets 
slightly more complicated.
}%

\commentout{
Again, it follows that the results of this paper and the previous paper
showing that \gseu\ can represent (almost) all \drules\ in the act
framework and all preference relations on acts apply to \drules\ and
preference relations in the horse lottery framework.
}%
}%
\commentout{
Since horse lotteries can be viewed as acts, it is immediate that the AA
framework is a special case of the act framework.
}%

\commentout{
Finally, note that $\E^{\utf,\widehat{\edom}}(\csq) = \top \widehat{\et} \utf(\csq) = \utf(\csq)$, 
so $\E^{\utf,\widehat{\edom}}$ is basically an extension of
$\utf$ to the roulette lotteries.

In fact, it is  quite common to let \utfn\ map
roulette lotteries to real numbers in the standard setting;
Of course, in the standard setting, it is typically assumed that 
the \prefrel\ on roulette lotteries satisfies some version of \vonNM's axioms
(otherwise there will not be a real-valued \utfn\ $\utf$ with the property that
$\utf(\lto) = \E^{\utf}(\lto)$); 
there is no need to do that here.

Also, in the standard setting, the \prefrel\ on roulette lotteries
}%

\section{Discussion}\label{sec:discussion}
We have shown that (almost) all \drules\ can be represented by \gseu\@.  
So what does this result buy us?  For one thing, \drules\ are typically
viewed as compact representations of how a DM makes decisions.
Our results suggest a uniform way of representing \drules\ that, in many
cases of interest, will be compact.  (How compact the representation is
depends on how compactly we can describe $\ep$, $\et$, and
$\eleq_{\eudom}$.  While natural choices for these functions and
relations typically do have a compact description, this is clearly not
the case for all possible choices.)  Our results also suggest a general
approach for constructing \drules.  This may be particularly relevant 
as we search for rules that are both adequate descriptively, in terms of
describing what people actually do, and computationally 
tractable.

\subsection*{Acknowledgments}  We thank Duncan Luce and Peter Wakker
for useful comments on a draft of the paper.

\appendix
\section{Proofs}

\commentout{
\othm{represent}  
For all \drules\ $\dr$, 
$\dr$ has a  \gseu\ \representation\ iff 
$\dr$ is \uniform\ and $\dr$ respects utility.
\eothm
}%
\setThm{repSecNum}{repThmNum}
\thm
For all \drules\ $\dr$, 
$\dr$ has a  \gseu\ \representation\ iff 
$\dr$ is \uniform\ and $\dr$ respects utility.
\ethm
\unsetThm
\prf 
\commentout{
Due to lack of space, we omit the 
argument that if $\dr$ is uniform and
respects utility, then it can be represented by 
\gseu\@.
``if'' direction.
}%
\commentout{
The ``if'' direction 
is
somewhat similar in spirit to the proof of Theorem~\ref{represent1},
given below;
\commentout{
Basically, we construct the \expstr\ (and \plmsr) so that 
$\eplut{\act} = \utf_\act$ in the 
\nonplstic\ case and $\eplut{\act} = \lto^{\plf,\utf}_\act$ in the \plstic\ 
case.  
}%
due to the lack of space, we omit this direction.
}%
We first show that if $\dr$ has a  \gseu\ \representation, then 
it is  \uniform\ and respects utility.
\commentout{
For the 
``only if'' direction, 
suppose 
}%
So, suppose
that $\trans$ is a \gseu\ \representation\ of $\dr$ and
let $\dprb_0 \in \dom(\dr)$ be arbitrary.
Suppose that 
$\act_1, \act_2, \actb_1, \actb_2$ are acts of $\dprb_0$ such that
$\act_i \ind_{\dprb_0} \actb_i$.  It is easy to check that
if $\dprb = (\ACT,\edom,\utf,\plf) \cong \dprb_0$, then 
$\eplut{\act_i} = \eplut{\actb_i}$.  Thus for all
\plstic\ $\dprb$, if 
$\dprb \cong \dprb_0$, then
\commentout{
$\act_1 \leqRPdp{\mr{\gseu}}{\dprb} \act_2$
 iff 
$\actb_1 \leqRPdp{\mr{\gseu}}{\dprb} \actb_2$. 
}%
\[
\act_1 \leqRPdp{\mr{\gseu}}{\dprb} \act_2
\mbox{ iff }
\actb_1 \leqRPdp{\mr{\gseu}}{\dprb} \actb_2. 
\]
Since $\trans$ is a \gseu\ \representation\ of $\dr$,
$\trans(\dprb_0) \cong \dprb_0$ and 
$\dr(\dprb_0) = \mr{\gseu}(\trans(\dprb_0))$.  It follows then
that
\commentout{
$\act_1 \leqRdp{\dprb_0} \act_2$
 iff 
$\actb_1 \leqRdp{\dprb_0} \actb_2$; 
}%
\[
\act_1 \leqRdp{\dprb_0} \act_2
\mbox{ iff }
\actb_1 \leqRdp{\dprb_0} \actb_2; 
\]
thus $\dr$ is \uniform.

Now suppose that $\act_1$ and $\act_2$ are two acts of constant utility, say
$u_1$ and $u_2$, respectively, of $\dprb_0$. Since 
$\trans(\dprb_0) \cong \dprb_0$, 
$\act_i$ is still an act of constant utility $u_i$ in $\trans(\dprb_0)$. 
Note that 
\commentout{
$\act_1 \leqRdp{\dprb_0} \act_2$ iff
$\act_1 \leqRPdp{\mr{\gseu}}{\trans(\dprb_0)} \act_2$ iff
$u_1 \eleq_{\udom} u_2$,
}%
\[
\act_1 \leqRdp{\dprb_0} \act_2 \mbox{ iff }
\act_1 \leqRPdp{\mr{\gseu}}{\trans(\dprb_0)} \act_2 \mbox{ iff }
u_1 \eleq_{\udom} u_2,
\]
where $\udom$ is the \utstr\ of $\dprb_0$,
since $\trans$ is a \gseu\ \representation\ of $\dr$.  Thus $\dr$ respects
utility.

We now show that, if $\dr$ is  \uniform\ and respects utility, then
it  has a  \gseu\ representation.
We begin with the \nonplstic\ case.

Suppose $\dr$ is a \uniform\ \nonplstic\ 
\drule\ that respects utility.
Fix 
some 
\dprob\
$\pdprb = ((\Act,\Stt,\Csq),\udom,\utf) \in \dom(\dr)$.
Let $\edom = (\udom,\pdom,\eudom,\ep,\et)$, where
$\pdom = (2^\Stt, \sbset)$, $\eudom = 2^{\Stt \times \udom}$,
$x \ep y = x \union y$, and 
$X \et u = X \times \{u\}$.  
Now define
$\eleq_\eudom$ 
as follows:
$x \eleq_\eudom y$ iff
\ben
\bl $x = y$, or
\bl 
\label{prf:drule1}
$x = \Stt \times \{u\}$ and $y = \Stt \times \{v\}$ for some $u, v \in \udom$ such
that $u \eleq_\udom v$, or
\bl 
\label{prf:drule2}
$x = \utf_{\act}$  and
$y = \utf_{\actb}$ for some $\act, \actb \in \Act$ such that
$\act \eleq_{\dr(\pdprb)} \actb$.  
\een
We need to check that $\eleq_\eudom$ is well defined.  
\commentout{
We also
need to 
}%
To see that \ref{prf:drule2} does not introduce any inconsistencies by itself,
we need to
show that whenever we have 
$\act_1, \actb_1, \act_2, \actb_2 \in \Act$   
such that 
$\utf_{\act_1} = \utf_{\act_2}$ and
$\utf_{\actb_1} = \utf_{\actb_2}$, then 
$\act_1 \eleq_{\dr(\pdprb)} \actb_1$ iff
$\act_2 \eleq_{\dr(\pdprb)} \actb_2$.
Here is where we use the assumption that $\dr$ is \uniform.
Note that
$\utf_{\act_1} = \utf_{\act_2}$ and 
$\utf_{\actb_1} = \utf_{\actb_2}$ implies that
$\act_1 \ind_{\pdprb} \act_2$ and 
$\actb_1 \ind_{\pdprb} \actb_2$.
Thus 
$\act_1 \eleq_{\dr(\pdprb)} \actb_1$ iff
$\act_2 \eleq_{\dr(\pdprb)} \actb_2$,
since $\dr$ is \uniform.
Note that \ref{prf:drule1} is essentially relating constant 
\urvs;
since $\dr$ respects utility, \ref{prf:drule1} and \ref{prf:drule2}
are consistent with one another.
Thus $\eleq_\eudom$ is well defined. 
We
identify $u \in \udom$ with $\Stt \times \{u\}$, so 
we have $\top \et u = u$, and it is clear that $\ep$ is associative and commutative.
Given \ref{prf:drule1}, it is easy to see that
$(\udom, \eleq_\udom)$ is a substructure of
$(\eudom, \eleq_\eudom)$.  
Thus, $\edom$ is an \expstr.

Let $\plf(X) = X$ 
and $\trans(\pdprb) = ((\Act,\Stt,\Csq),\edom,\utf,\plf)$.
It is clear that $\trans(\pdprb) \cong \pdprb$, since
the \dsitu\ and \utfn\ have not changed.  Given the definitions of
$\eplut{\act}$, $\edom$, $\plf$, and $\utf$, we have
\[
\begin{array}{lcl}
\eplut{\act} 
& = & \ds \EP_{u \in \ran(\utf_\act)} \plf(\utf_{\act}^{-1}(u)) \et u \\
& = & \ds \bigcup_{u \in \ran(\utf_\act)} \utf_{\act}^{-1}(u) \times \{ u \} \\
& = & \ds \{ (s,u) \st 
u \in \ran(\utf_\act)
\mbox{ and } 
s \in \utf_{\act}^{-1}(u) 
\}\\
& = & \utf_\act.
\end{array}
\]
Given the definition of $\eleq_\eudom$ and the 
fact that $\eplut{\act} =  \utf_\act$ for all $\act \in \Act$,
it is
immediate that $\mr{\gseu}(\trans(\pdprb)) = \dr(\pdprb)$.
Thus $\trans$ is a \gseu\ \representation\ of $\dr$. 

The argument for the \plstic\ case is 
completely analogous,
so we give a sketch here and leave the details to the reader.
The key difference is that,
instead of having $\pdom = (2^\Stt, \sbset)$ and $\plf(X) = X$, the \plstr\ and
\plmsr\ are already givens.
So, 
instead of making
$\eplut{\act} = \utf_\act$ 
(which is not possible in general, since we have to
use the given \plmsr), 
we make
$\eplut{\act} = \lto^{\plf,\utf}_{\act}$;
that is, $\eplut{\act}$ is 
the \ulottery\
induced by $\act$ instead of the \urv\
induced by $\act$.  

Suppose $\dr$ is a \uniform\ \plstic\ \drule\ that respects utility.
Fix
some 
\plstic\ \dprob\ 
$\pdprb = ((\Act,\Stt,\Csq),\edom_1,\utf,\plf) \in \dom(\dr)$.
Let $\edom_2 = (\udom_1,\pdom_1,\eudom,\ep,\et)$, where
$\udom_1$ is the \utstr\ of $\edom_1$,
$\pdom_1$ is the \plstr\ of $\edom_1$,
$\eudom = 2^{\pdom_1 \times \udom_1}$,
$x \ep y = x \union y$, and 
$p \et u = \{(p, u)\}$.  
Define $\eleq_{\eudom}$ as follows:
$x \eleq_{\eudom} y$ iff
\ben
\bl $x = y$, or
\bl 
\label{prf:drule2-1}
$x = \{(\top,u)\}$ and $y = \{(\top,v)\}$ for some $u, v \in \udom_1$
such that $u \eleq_{\udom_1} v$, or
\bl 
\label{prf:drule2-2}
$x = \lto^{\plf,\utf}_\act$ and 
$y = \lto^{\plf,\utf}_\actb$ for some $\act, \actb \in \Act$ such that
$\act \eleq_{\dr(\pdprb)} \actb$.  
\een
Again, we need to check that $\eleq_\eudom$ is well defined.
As in the \nonplstic\ case,  it is easy to check that
\ref{prf:drule2-2} does not introduce inconsistencies by itself, since $\dr$ is
\uniform. 
Also, 
since $\dr$ respects utility, \ref{prf:drule2-1} and \ref{prf:drule2-2}
are consistent with one another.
\commentout{
We also
need to show that whenever we have 
$\act_1, \actb_1, \act_2, \actb_2 \in \Act$   
such that 
$\lto^{\plf,\utf}_{\act_1} = \lto^{\plf,\utf}_{\act_2}$ and
$\lto^{\plf,\utf}_{\actb_1} = \lto^{\plf,\utf}_{\actb_2}$, then
$\act_1 \eleq_{\dr(\dprb)} \actb_1$ iff
$\act_2 \eleq_{\dr(\dprb)} \actb_2$.
Here is where we use the assumption that $\dr$ is \uniform.
Note that if $\lto^{\plf,\utf}_{\act_1} = \lto^{\plf,\utf}_{\act_2}$ and
$\lto^{\plf,\utf}_{\actb_1} = \lto^{\plf,\utf}_{\actb_2}$, then
$\act_1 \ind_\dprb \act_2$ and 
$\actb_1 \ind_\dprb \actb_2$.  Since $\dr$ is \uniform, we indeed have
$\act_1 \eleq_{\dr(\dprb)} \actb_1$ iff
$\act_2 \eleq_{\dr(\dprb)} \actb_2$.
}%
We identify $u \in \udom$ with $\{(\top,u)\}$,
so $\top \et u = u$;
given \ref{prf:drule2-1},
it is easy to see that
$(\udom,\eleq_\udom)$ is a substructure of $(\eudom,\eleq_\eudom)$. 
Again, $\ep$ is associative and commutative.
Thus $\edom_2$ is an \expstr.

Let $\trans(\pdprb) = (\ACT,\edom_2,\utf,\plf)$.
Obviously, $\trans(\pdprb) \cong \pdprb$, since the \dsitu, \utfn, and \plmsr\
have not changed.
It is easy to verify that 
$\ePplut{\edom_2}{\act} = \lto^{\plf,\utf}_\act$
for all $\act \in \Act$.
Thus it is immediate that
$\mr{\gseu}(\trans(\pdprb)) = \dr(\pdprb)$, given the definition of
$\eleq_\eudom$, so
$\trans$ is a \gseu\ \representation\ of $\dr$.  
\eprf

\commentout{
\othm{represent1}  
A \drule\ $\dr$ has an ordinal \gseu\ \representation\ 
iff $\dr$ weakly respects utility.
\eothm
}%
\setThm{repoSecNum}{repoThmNum}
\thm
A \drule\ $\dr$ has an ordinal \gseu\ \representation\ 
iff $\dr$ weakly respects utility.
\ethm
\unsetThm
\prf
We first show that if $\dr$ has an ordinal  \gseu\ \representation, then 
it is  weakly respects utility.
So, Suppose that $\trans$ is an ordinal \gseu\ \representation\ of  $\dr$.  
Let $\dprb_1 \in \dom(\dr)$ be arbitrary. 
Suppose that $\act_{\csq_1}$
$\act_{\csq_2}$ are constant acts in $\dprb_1$ (where 
$\act_{\csq_i}(\stt) = \csq_i$ for all  states $\stt$). We need to show that 
\[
\act_{\csq_1} \leqRdp{\dprb_1} \act_{\csq_2} \mbox{ iff }
\utf_1(\csq_1) \eleq_{\udom_1} \utf_1(\csq_2), 
\]
where $\utf_1$ is the \utfn\ of $\dprb_1$
and $\udom_1$ is the \utstr\ of $\dprb_1$. 
Let $\dprb_2 = \trans(\dprb_1)$;
since $\trans$ is an ordinal \gseu\ \representation\ of $\dr$, 
$\dprb_2 \ssim \dprb_1$ and 
$\mr{\gseu}(\dprb_2) = \dr(\dprb_1)$. So
\[
\act_{\csq_1} \leqRdp{\dprb_1} \act_{\csq_2} \mbox{ iff }
\act_{\csq_1} \leqRPdp{\mr{\gseu}}{\dprb_2} \act_{\csq_2} \mbox{ iff }
\utf_2(\csq_1) \eleq_{\udom_2} \utf_2(\csq_2),
\]
where $\utf_2$ is the \utfn\ of $\dprb_2$ and $\udom_2$ is the \utstr\ of $\dprb_2$.
Since $\dprb_2 \ssim \dprb_1$,
\[
\utf_2(\csq_1) \eleq_{\udom_2} \utf_2(\csq_2) \mbox{ iff } 
\utf_1(\csq_1) \eleq_{\udom_1} \utf_1(\csq_2), 
\]
and we see that $\dr$ weakly respects utility.

Now we show that if $\dr$ weakly respects utility, then it has an ordinal
\gseu\ \representation.
As in \mthm{represent}, there are two cases, \plstic\ and \nonplstic.  
They are almost identical, so we do just
the \plstic\ case here. 
\commentout{
(Also, the ``only if'' direction is 
quite similar to the one in the proof of \mthm{represent}, so we omit it here.)
}%

Suppose that $\dr$ is a \plstic\ \drule\
that weakly respects utility.
 Fix a \plstic\ \dprob\
$\dprb = ((\Act,\Stt,\Csq), \edom_1, \utf_1, \plf_1) \in \dom(\dr)$.
\commentout{
We need
to construct a  \dprob\ $\dprb_2 = ((\Act,\Stt,\Csq), \edom_2, \utf_2,\plf_2)$ such that
$\dprb_2 \ssim \dprb_1$ and
$\mr{\gseu}(\dprb_2) = \dr(\dprb_1)$.
}%
Let $\udom_1$ and $\pdom_1$ be the \utstr\ and \plstr\ of $\edom_1$,
respectively.  Let $\edom_2 = (\udom_2,\pdom_2,\eudom,\ep,\et)$ be
defined as 
follows (note that $\times$ denotes Cartesian product in this proof):
\commentout{
\bit
\bl 
$\udom_2 = \udom_1 \times 2^\Csq$ and for all function symbols 
$f$ in the signature of $\udom_1$,
\[
f_{\udom_2}((u_1,X_2), (u_2,X_2)) = (f_{\udom_1}(u_1,
 u_2), X_1 \cup X_2)
\]
and for all relation symbols $R$ 
in the signature of $\udom_1$,
\[
((u_1, X_1), (u_2,X_2)) \in R_{\udom_2} \mbox{ iff } 
(u_1,  u_2) \in R_{\udom_1};\footnote{Due to the lack of space, we assume that
all functions and relations are binary; it is easy to generalize to $n$-ary cases.}
\]
note that, in particular, 
\[
(u_1,X_1) \eleq_{\udom_2} (u_2,X_2) \mbox{ iff } u_1
\eleq_{\udom_1} u_2.
\]
\eit
}%
\bit
\bl $\udom_2 = (\udom_1 \times \Csq, \eleq_{\udom_2})$, where
$(u_1,\csq_1) \eleq_{\udom_2} (u_2,\csq_2)$ iff $u_1 \eleq_{\udom_1} u_2$.
\bl $\pdom_2 = (\pdom_1 \times 2^\Stt, \eleq_{\pdom_2})$, where
$(p_1,X_1) \eleq_{\pdom_2} (p_2,X_2)$ iff $p_1 \eleq_{\pdom_1} p_2$.
(Note that $\eleq_{\pdom_2}$ is a partial preorder, although it is not a
partial order.)
\bl
$\eudom = (2^{\Stt \times \udom_2},\eleq_\eudom)$, where
$x \eleq_\eudom y$ iff $x = y$ or 
\ben
\bl
\label{prf:exp1}
$x = \Stt \times \{(u_1,c_1)\}$, $y = \Stt \times \{(u_2,c_2)\}$, and 
$(u_1,c_1) \eleq_{\udom_2} (u_2,c_2)$, or 
\bl
\label{prf:exp2}
$x = \{(\stt, (\utf_1(\act(\stt)), \act(\stt))) \st \stt \in \Stt\}$,
$y = \{(\stt, (\utf_1(\actb(\stt)), \actb(\stt))) \st \stt \in \Stt\}$, and 
$\act \eleq_{\dr(\dprb)} \actb$, for some $\act, \actb \in \Act$.
\een
\bl
$(p,X) \et (u,\csq) = X \times \{(u,\csq)\}$. 
\bl
$x \ep y = x \cup y$ for all $x,y \in \eudom$.
\eit
Note  that 
$(\bottom_{\pdom_1},\eset) \eleq_{\pdom_2} (p,X) \eleq_{\pdom_2} (\top_{\pdom_1},\Stt)$,  
so we have
$\bottom_{\pdom_2} = (\bottom_{\pdom_1},\eset)$ and
$\top_{\pdom_2} = (\top_{\pdom_1},\Stt)$;
thus, $\pdom_2$ is a \plstr. 
Since $\dr$ weakly respects utility,  
\ref{prf:exp1} and \ref{prf:exp2} are consistent with one another. 
We identify $(u,\csq) \in \udom_2$ with $\Stt \times \{(u,\csq)\}$ in $\eudom$;
with this identification, 
\commentout{
$(\udom_2, \eleq_{\udom_2})$ is a substructure of
$(\eudom,\eleq_\eudom)$ and $\top \et (u,\csq) = (u,\csq)$ for all 
$(u,\csq) \in \udom_2$ as required.
}%
$\top \et (u,\csq) = (u,\csq)$ for all 
$(u,\csq) \in \udom_2$ and, given \ref{prf:exp1} in the definition of
$\eleq_\eudom$, 
$(\udom_2, \eleq_{\udom_2})$ is a substructure of
$(\eudom,\eleq_\eudom)$.
Furthermore, $\ep$ is clearly associative and commutative, so $\edom_2$ is indeed
an \expstr.

Now 
we need to define a \utfn\ and a \plmsr.
Let $\utf_2(\csq) = (\utf_1(\csq),\csq)$ for all $\csq \in \Csq$ and let
$\plf_2(X) = (\plf_1(X),X)$ for all $X \sbset \Stt$.
\commentout{
To see that $\dprb_2$ is 
a \dprob, we need
to check that $\plf_2$ is a \plmsr.  
}%
Note that 
\beqn
\label{eqn:plford}
\plf_2(X) \eleq_{\pdom_2} \plf_2(Y) \mbox{ iff } \plf_1(X) \eleq_{\pdom_1} \plf_1(Y).
\eeqn
Thus $\plf_2$ is a \plmsr, since
$\plf_1$ is a \plmsr.
Also,
\beqn
\label{eqn:utord}
\utf_2(\csq) \eleq_{\udom_2} \utf_2(\csqd) \mbox{ iff }
\utf_1(\csq) \eleq_{\udom_1} \utf_1(\csqd).  
\eeqn
Let $\trans(\dprb) = ((\Act,\Stt,\Csq),\edom_2,\utf_2,\plf_2)$.
\commentout{
This fact combined with the 
observation above
about $\plf_2$ shows that 
$\dprb_2 \ssim \dprb_1$.  
}%
Note that, by \meqn{eqn:plford} and \meqn{eqn:utord}, $\trans(\dprb)
\ssim \dprb$; 
furthermore, 
it is easy to check that 
$\E_{\plf_2,\edom_2}((\utf_2)_{\act}) = \{ (\stt, (\utf_1(\act(\stt)),\act(\stt))) \st \stt \in \Stt\}$;
so 
$\mr{\gseu}(\trans(\dprb)) = \dr(\dprb)$,
given the definition of $\eleq_\eudom$.
Thus $\trans$ is an ordinal \gseu\ \representation\ of $\dr$.
\eprf

\commentout{
\opro{pro:stt-induce-lto2}
Every lottery \dsitu\ $\LTO = (\Lto,\Csq,\pdom)$ is induced by some 
\plstic\ \dsitu\ 
$\STT_\LTO$.
\eopro
}%
\setThm{redSecNum}{redThmNum}
\pro
Every lottery \dsitu\ $\LTO = (\Lto,\Csq,\pdom)$ is induced by some 
\plstic\ \dsitu\ 
$\STT_\LTO$.
\epro
\unsetThm
\prf
We first prove the proposition for the standard case.
Suppose that $\LTO = (\Lto,\Csq,[0,1])$.
Let $\Stt = [0,1)$.
Suppose that $\lto \in \Lto$ and $\supp(\lto) = \{c^\lto_1, \ldots, c^\lto_k\}$.
Let $\act_\lto$ be  defined as follows:
$\act_\lto(\stt) = \csq^\lto_k$ for all  
$s \in \Stt$ such that
$\sum_{i=1}^{k-1} \lto(\csq^\lto_i) \leq \stt < \sum_{i=1}^{k} \lto(\csq^\lto_i)$.
Let $\STT_\LTO = ((\Act_\Lto,\Stt,\Csq),[0,1],\Pr)$, 
where
\bit
\bl 
$\Pr$ is the uniform distribution on $\Stt$ and
\bl 
$\Act_\Lto = \{ \act_\lto \st \lto \in \Lto\}$. 
\eit
It is easy to check that 
$\lto^{\Pr}_{\act_\lto} = \lto$, 
so $\STT_\LTO$
induces $\LTO$. 

The construction is more complicated for general \plstrs, since we
must make sure $\Stt$ is rich enough 
to allow us to use a single \plmsr\ to induce
all the lotteries.
Given a lottery \dsitu\ $\LTO = (\Lto,\Csq,\pdom)$, let 
$\Stt_\Lto = \{ f  \st f \in \Csq^\Lto$ and $f(\lto) \in \supp(\lto) \}$.
Intuitively, each state 
$f$ assigns to each lottery $\lto$ some consequence in $\supp(\lto)$.
Let $\act_\lto$ be defined by taking $\act_\lto(f) = f(\lto)$.  
\commentout{
\bl
$\plf$ is a \plmsr\ on $\Stt$ such that
$\plf(\{ f \st f(\lto) = i\}) = \lto(\csq_i)$.\footnote{Such \plmsrs\ exist
since for all $\lto_1, \lto_2 \in \Lto$,  
there is no subset relation between 
$\{ f \st f(\lto_1) = i\}$ and $\{ f \st f(\lto_2) = j \}$, so we can assign
any plausibility value we want to these sets.}
}%
Now we need to specify a \plmsr.  
The idea is to 
construct $\plf$ so that
$\plf(\act^{-1}_{\lto}(X)) = \lto(X)$.
Clearly this 
guarantees that 
$\lto^\plf_{\act_\lto}(X) = \lto(X)$ for all $X \in 2^{\supp(\lto)}$,
so that $\act_\lto$ induces
$\lto$.  
\commentout{
However, we must make sure that, with this definition, $\plf$
is a well defined \plmsr, and that it can be extended to all of $2^S$.
Formally, given $Y \sbset \Stt_\Lto$, we define $\plf(Y)$ as follows:
\ben
\bl if there exists $X \subseteq X' \subseteq \Csq$ and $\lto \in \Lto$
such that 
\begin{itemize}
\item[(a)] $X' \neq \dom(\lto)$;
\item[(b)] $\plf(\act^{-1}_{\lto}(X)) \subseteq Y \subseteq
\plf(\act^{-1}_{\lto}(X'))$; 
\item[(c)] there does not exist $X''$ such that either  $X \subset X''
\subseteq X'$ and $\plf(\act^{-1}_{\lto}(X'')) \subseteq Y \subseteq
\plf(\act^{-1}_{\lto}(X'))$ or $X \subseteq X'' \subset X'$ 
and $\plf(\act^{-1}_{\lto}(X)) \subseteq Y \subseteq
\plf(\act^{-1}_{\lto}(X''))$ (where $\subset$ denotes strict subset)
\end{itemize}
then $\plf(Y) = \lto(X)$;
\bl otherwise, $\plf(Y) = \top$.
\een
}%

To make the definition of $\plf$ more concise, let $\vp(\lto,Y)$ be the
following statement: there exists some nonempty $X \sbset \supp(\lto)$ such
that $\act^{-1}_\lto(X) \sbset Y$.
Given $Y \sbset \Stt_\Lto$, we define $\plf(Y)$ as follows:
\ben
\bl
If there does not exist $\lto \in \Lto$ such that 
\commentout{
there exists some nonempty 
$X \sbset \supp(\lto)$ 
such that $\act^{-1}_{\lto}(X) \sbset Y$, 
}%
$\vp(\lto,Y)$, 
let $\plf(Y) = \bottom$.
\bl
\label{tempref}
If there exists a unique $\lto \in \Lto$ such that
$\vp(\lto,Y)$, 
let $\plf(Y) = \lto(Z)$, where
\[
Z = \bigcup \{ X \st X \sbset \supp(\lto) \mbox{ and }\act^{-1}_\lto(X) \sbset Y\}.
\]
\bl
\commentout{
If none of the above applies 
(so, roughly speaking, there exist two distinct
elements of $\Lto$ 
that satisfy the condition in Case \ref{tempref}), 
}%
If there exist two distinct $\lto_1, \lto_2 \in \Lto$ such that
$\vp(\lto_1,Y)$ and $\vp(\lto_2,Y)$, 
let $\plf(Y) = \top$. 
\een
\commentout{
We need to verify that $\plf$ is indeed a \plmsr.  It is clear that
$\plf(\eset) = \bottom$, since 
$\act^{-1}_\lto(\csq) \neq \eset$ for $\csq \in \supp(\lto)$, so case 1
applies. 
To see that 
$\plf(\Stt_\Lto) = \top$, we have two cases:
\bit
\commentout{
\bl $\abs{\Lto} = 1$.  In this case, case 2 applies and it is clear that 
$Z^* = N(\lto)$ for the sole $\lto \in \Lto$. Since $\lto$ is a lottery,
$\plf(\Stt_\Lto) = \lto(\supp(\lto)) = \top$.
\bl $\abs{\Lto} > 1$.  In this case, case 3 applies and it is immediate that
$\plf(\Stt_\Lto) = \top$.
}%
\bl
There exists some degenerate lottery $\lto \in \Lto$.  Then
$\Stt_\Lto = \act^{-1}_{\lto}(\csq)$ for the sole member of $\supp(\lto)$. 
Since $\lto$ satisfies \ref{lto:req1}, 
$\plf(\Stt_\Lto) = \lto(\csq) = \top$.
\bl
There exists some nondegenerate lottery $\lto \in \Lto$.  Then
$\Stt_\Lto$ is a proper superset of $\act^{-1}_{\lto}(\csq)$ for any
$\csq \in \supp(\lto)$, so case 3 applies.
\eit
Since $\Lto \neq \eset$, one (or both) of the above applies.  
Now we need to check that if $Z_1 \sbset Z_2 \sbset \Stt_\Lto$, then 
$\plf(Z_1) \pleq \plf(Z_2)$.  We 
have three cases:
\bit
\bl 
Case 1 applies to $Z_2$.  Then it clearly applies to $Z_1$ as well, so 
$\plf(Z_1) = \bottom \pleq \plf(Z_2)$.
\bl
Case 2 applies to $Z_2$.  If case 1 applies to $Z_1$, then we are done as above.
Otherwise, it is clear that $Z_1 = Z_2$, so $\plf(Z_1) = \plf(Z_2)$.
\bl
Case 3 applies to $Z_2$.  Then $\plf(Z_1) \pleq \top = \plf(Z_2)$. 
\commentout{
\bl Case 1 applies to $T_2$.  Then it clearly applies to $T_1$ as well, so
$\plf(T_1) = \bottom \pleq \plf(T_2)$.
\bl Case 2 applies to $T_2$.  If case 1 applies to $T_1$, then we are done as
above. If case 2 applies to $T_1$ as well, it is clear that $T_1$ shares the
same lottery $\lto$ with $T_2$ and $Z^*_1 \sbset Z^*_2$
(where $Z^*_i$ is the set associated with $T_i$).   Thus
$\plf(T_1) = \lto(\{ \csq^{\lto}_i \st i \in Z^*_1 \})
\pleq \lto(\{ \csq^{\lto}_i \st i \in Z^*_2 \}) = \plf(T_2)$, since $\lto$ is a
lottery.
\bl Case 3 applies to $T_2$.  Then $\plf(T_1) \pleq \top = \plf(T_2)$.
}%

\eit
So $\plf$ is indeed a \plmsr.
It is easy to check that 
$\lto^\plf_{\act_\lto} = \lto$,
so $\STT_\LTO$ indeed indcues $\LTO$.
}%
\commentout{
We need to verify that $\plf$ is well defined and a \plmsr.  
To see that it is well defined, suppose that 
there exist $X_1 \subseteq X_1' \subseteq \Csq$ and $\lto_1 \in \Lto$
and $X_2 \subseteq X_2' \subseteq \Csq$ and $\lto_2 \in \Lto$ satisfying
the first clause in the definition with respect to $Y$ and $\plf(X_1)
\ne \plf(X_2)$.  If $\lto_1 = \lto_2$, then we must have $X_1 = X_2$,
for otherwise $X_1 \subset X_1 \union X_2$ and 
$\plf(\act^{-1}_{\lto}(X_1 \union X_2)) \subseteq Y$, contradicting the
choice of $X_1$.  If $\lto_1 \ne \lto_2$, note that
at least one of $X_1$ and $X_2$ is nonempty.
Suppose without loss of generality that $X_1 \ne \emptyset$  and that
$\csq \in \dom(\lto_2) - X_2'$.  Choose $f \in
C^L$ such that $f(\lto_1) \in X_1$ and $f(\lto_2) = \csq$.  Then clearly
$f \in \act^{-1}_{\lto_1}(X_1) - \act^{-1}_{\lto_2}(X_2')$.  However, since
$\act^{-1}_{\lto_1}(X_1) \subseteq Y$ and $Y \subseteq
\act^{-1}_{\lto_2}(X_2)$, we must have $\act^{-1}_{\lto_1}(X_1)
\subseteq \act^{-1}_{\lto_2}(X_2)$.  This gives a contradiction.
Thus, $\plf$ is well defined.  
}%
Note that 
for
each $Y \sbset \Stt_\Lto$, 
exactly one of the three cases applies,
so $\plf$ is well defined.

To see that $\plf$ is a plausibility measure, note that
clearly $\plf(\Stt_\Lto) = \top$ 
(since $\Lto \neq \eset$)
and $\plf(\emptyset) = \bot$.
Now suppose that $Y_1 \subseteq Y_2$.  
\commentout{
If $\plf(Y_2) = \top$, then
clearly $\plf(Y_1) \pleq \plf(Y_2)$.  Otherwise, it must be the case
that there exist $X, X', \lto$ such that case 1 of the definition of
$\plf$ applies and $\plf(Y_2) = \lto(X)$.  Since $Y_1 \subseteq Y_2$, 
it is immediate that there
exists $X_1, X_1'$ such that $X_1 \subseteq X$ and $X_1' \subseteq X'$
such that case 1 of the definition applies to $X_1, X_1, \lto$ with
respect to $Y_1$.  Thus, $\plf(Y_1) = \lto(X_1) \pleq \plf(Y_2)$.  
}%
We have three cases:
\bit
\bl Case 1 applies to $Y_2$.  Then it must apply to $Y_1$ as well, so
$\plf(Y_1) = \bottom \pleq \plf(Y_2)$.
\bl Case 2 applies to $Y_2$; let $\lto_2$ be the unique lottery 
such that $\vp(\lto_2, Y_2)$.
Since $Y_1 \sbset Y_2$, 
for all $\lto \in \Lto$,
$\vp(\lto,Y_1)$ implies $\vp(\lto,Y_2)$.
Thus, if there is some $\lto \in \Lto$
such that 
$\vp(\lto,Y_1)$, it must be $\lto_2$.
So
either
case 1 applies to $Y_1$, then we are done as
above, or
$\vp(\lto_2,Y_1)$.
\commentout{
lottery associated with $Y_1$.  
unique lottery such that $\vp(\lto_1, Y_1)$.
Note that if $X$ is a nonempty subset such that 
$\act^{-1}_{\lto_1}(X) \subseteq Y_1$, then 
$\act^{-1}_{\lto_1}(X) \subseteq Y_2$.  By the uniqueness of $\lto_2$, it
follows that 
$\lto_1 = \lto_2$.
}%
Since $Y_1 \sbset Y_2$, if $\act^{-1}_{\lto_2}(X) \sbset Y_1$ then 
$\act^{-1}_{\lto_2}(X) \sbset Y_2$; thus $Z_1 \sbset Z_2$, where 
\[
Z_i = \bigcup \{ X \st X \sbset \dom(\lto_2)\mbox{ and }
\act^{-1}_{\lto_2}(X) \sbset Y_i\},
\]
and so $\plf(Y_1) = \lto_2(Z_1) \pleq \lto_2(Z_2) = \plf(Y_2)$.
\bl
Case 3 applies to $Y_2$.  Then $\plf(Y_1) \pleq \top = \plf(Y_2)$.
\eit
So $\plf$ is a \plmsr.
\commentout{
Moreover, $\plf(\act^{-1}_{\lto}(X)) = \lto(X)$.  Clearly this is true
if $X = \emptyset$ or $X = \supp(\lto)$.  So suppose that $X$ is a
nonempty strict subset of $\supp(\lto)$.  The question is whether case 2
or case 3 of the definition of $\plf$ applies in computing
$\plf(\act^{-1}_{\lto}(X))$.  If it is case 2, then we are done.  We now
show that it cannot be case 3.  For suppose there exists 
a lottery $\lto'$ and a nonempty $X' \subseteq \supp(\lto')$ such that
$\act^{-1}_{\lto'}(X') \subseteq \act^{-1}_{\lto}(X)$.  Since $X \ne
\supp(\lto)$, choose some $c \in \supp(\lto) - X$ and some $c' \in X'$.
Choose some $f \in \Stt_\Lto$ such that $f(\lto) = c$ and $f(\lto') = c'$.
Then $f = \act^{-1}_{\lto'}(X') - \act^{-1}_{\lto}(X)$.  Thus case 2
must apply.  
It follows
that $\STT_\LTO$ induces $\LTO$. 
}%

Now we want to show that $\plf(\act^{-1}_{\lto}(X)) = \lto(X)$ for all 
$X \sbset \supp(\lto)$.  Clearly this is true
if $X = \emptyset$ or $X = \supp(\lto)$.  
So suppose that $X$ is a \npsub\ of $\supp(\lto)$.
Note that $\vp(\lto, \act^{-1}_{\lto}(X))$, so either case 2 or case 3 of the
definition of $\plf$ applies.  
Suppose that
$\vp(\lto_0, \act^{-1}_{\lto}(X))$ for some $\lto_0 \in \Lto$.  
Then there exists some nonempty $X_0 \sbset \supp(\lto_0)$ 
such that $\act^{-1}_{\lto_0}(X_0) \sbset \act^{-1}_\lto(X)$.
We want to show that $\lto_0 = \lto$, so that case 2 applies.
Note that 
there exists some $\csq \in \supp(\lto) - X$ and there exists
some $\csq_0 \in X_0$ by assumption.  Suppose that $\lto_0 \neq \lto$; then
there exists 
some 
$f \in \Stt_\Lto$ such that $f(\lto) = \csq$ and 
$f(\lto_0) = \csq_0$ by construction.  However, it is clear that 
$f \in \act^{-1}_{\lto_0}(X_0)$ and $f \notin \act^{-1}_{\lto}(X)$. Since
$\act^{-1}_{\lto_0}(X_0) \sbset \act^{-1}_\lto(X)$, no such $f$ exists; it
follows that $\lto_0 = \lto$ and so case 2 applies.
Thus
$\plf(\act^{-1}_{\lto}(X)) = \lto(X)$ and so
$\STT_\LTO$ induces $\LTO$. 
\eprf

\bibliographystyle{chicago}
\bibliography{z,joe,refs,newutil}
\end{document}